\documentclass{article}
\usepackage{iclr2027_conference,times}

\usepackage[utf8]{inputenc}
\usepackage[T1]{fontenc}
\usepackage{url}
\usepackage{booktabs}
\usepackage{amsfonts}
\usepackage{nicefrac}
\usepackage{microtype}
\usepackage[table]{xcolor}
\usepackage{multirow}
\usepackage{makecell}
\usepackage{tikz}
\usepackage{pgfplots}
\usepackage{multicol}
\usepackage{amsmath}
\usepackage{wrapfig}
\usepackage{algorithm}
\usepackage{algpseudocode}
\usepackage{amssymb}
\usepackage{enumitem}
\usepackage{caption}
 \usepackage{booktabs,tabularx}
\usepackage[most]{tcolorbox}

\definecolor{lavender}{HTML}{E6E6FA}

\newtcolorbox{samplebox}[1]{
	enhanced,
	colback=teal!10!white,
	colframe=black!60,
	boxrule=0.6pt,
	arc=6pt,
	left=3pt, right=3pt, top=2pt, bottom=2pt,
	boxsep=1.2pt,
	width=1.1\linewidth,
	before upper={\noindent\textbf{\small #1}\par\smallskip},
}
\usepackage[pagebackref,breaklinks,colorlinks]{hyperref}
\hypersetup{
   colorlinks=true,
   linkcolor=teal,
   citecolor=teal,
   urlcolor=teal
}
\title{Self-Regularized Flow Language Models}

\author{Theodoros Kouzelis$^{1,2}$\thanks{Correspondence to \texttt{theodoros.kouzelis@athenarc.gr}}, Spyros Gidaris$^{3}$, Nikos Komodakis$^{1,4,5}$ \\
$^{1}$Archimedes, Athena RC \quad $^{2}$National Technical University of Athens \\
$^{3}$valeo.ai \quad $^{4}$University of Crete \quad $^{5}$IACM-Forth
}

\newcommand{\Th}[1]{\textsc{#1}}

\newcommand{\red}[1]{{\textcolor{red}{#1}}}

\newcommand{\citeme}[1]{\red{[XX]}}
\newcommand{\refme}[1]{\red{(XX)}}

\makeatletter
\newcommand*\bdot{\mathpalette\bdot@{.7}}
\newcommand*\bdot@[2]{\mathbin{\vcenter{\hbox{\scalebox{#2}{$\m@th#1\bullet$}}}}}
\makeatother

\makeatletter
\DeclareRobustCommand\onedot{\futurelet\@let@token\@onedot}
\def\@onedot{\ifx\@let@token.\else.\null\fi\xspace}

\makeatother

\definecolor{ForestGreen}{RGB}{34,139,34}
\definecolor{Orange}{RGB}{1.0, 0.55, 0.0}
\definecolor{Purple}{RGB}{0.58, 0.44, 0.86}
\definecolor{softblue}{rgb}{0.2, 0.4, 0.8}
\definecolor{lightblue}{rgb}{0.6, 0.75, 0.95}
\definecolor{darkblue}{rgb}{0.05, 0.2, 0.5}

\definecolor{coral}{RGB}{255, 127, 80}
\definecolor{softcoral}{rgb}{1.00, 0.60, 0.50}

\definecolor{teal}{rgb}{0.1, 0.6, 0.6}
\definecolor{forestgreen}{rgb}{0.1, 0.6, 0.2}

\definecolor{vibrantorange}{rgb}{1.0, 0.5, 0.0}

\definecolor{deepblue}{rgb}{0.0, 0.0, 0.8}

\definecolor{TableColor}{rgb}{0.58, 0.55, 0.51}

\definecolor{TableColor}{rgb}{0.92, 0.95, 0.99}
\definecolor{TableColorGrey}{rgb}{0.88, 0.88, 0.88}

\definecolor{DarkenedMagenta}{RGB}{225,0,100}

\newcommand{\Eq}[1]{\hyperref[#1]{Eq.~(\ref{#1})}}
\newcommand{\Equation}[1]{\hyperref[#1]{Equation~(\ref{#1})}}
\newcommand{\secref}[1]{\hyperref[#1]{Sec.~\ref{#1}}}

\iclrfinalcopy
\begin{document}

\maketitle

\begin{abstract}
We introduce \texttt{SERF} (\textbf{Se}\textit{lf}-\textbf{R}\textit{egularized} \textbf{F}\textit{low Language Models}), a simple \textit{self-supervised} framework for improving continuous diffusion language models by exploiting the denoising trajectory itself. While continuous flow-based language models offer a promising alternative to autoregressive generation, their training objectives typically rely on supervision from clean target tokens, leaving the denoising trajectory underexploited as a learning signal. \texttt{SERF} addresses this limitation through noisy-to-clean self-distillation: an exponential moving average (EMA) teacher receives a cleaner version of the student's noisy input along the same interpolation path and provides additional supervision. We investigate two variants: \texttt{SERF-L}, which aligns the student's intermediate representations with deeper teacher representations, and \texttt{SERF-P}, which aligns their output token distributions, either at the student's output or at an early student layer. All variants augment the standard training objective without requiring external pretrained models or modifying the inference procedure. Experiments across three continuous language-model paradigms (one-hot \texttt{FLM}, latent \texttt{ELF}, and hyper-spherical \texttt{S-FLM}) demonstrate substantial improvements in training efficiency, generation quality, and few-step generation. In particular, \texttt{SERF} achieves up to $7.7\times$ wall-clock training acceleration on OpenWebText, improves MAUVE by $24.3\%$ and GSM8K accuracy by up to $28.8\%$, and matches baseline performance with up to $64\times$ fewer sampling steps. These results establish noisy-to-clean self-distillation as an effective approach to improving continuous diffusion language models.   \\ 
Code: \url{https://github.com/zelaki/SERF}
\end{abstract}

\begin{figure*}[h]
\vspace{-1pt}
  \centering
\includegraphics[width=\textwidth]{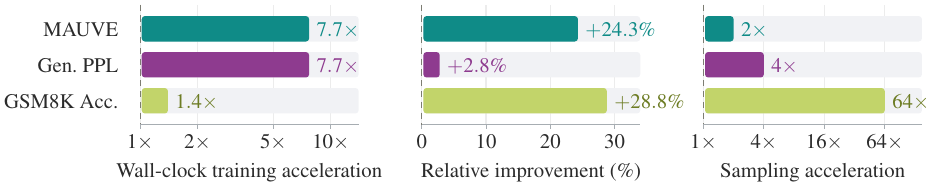}
  \vspace{-10pt}
\caption{\textbf{Performance and efficiency gains of \texttt{SERF}.} Improvements over
\texttt{FLM} ~\citep{lee2026flow} on \texttt{OpenWebText} ~\citep{gokaslan2019openwebtext} and over
\texttt{S-FLM}~\citep{deschenaux2026language} on \texttt{GSM8K}.
\textbf{Left:} SERF reaches the baseline's quality with substantially fewer training steps (see \autoref{fig:various}), achieving wall-clock acceleration despite its small additional computational cost (see \autoref{sec:additiona_serf_details}).
\textbf{Middle:}
 Improves generation quality and reasoning accuracy at the same
sampling budget (see \autoref{tab:best_mauve_steps} and \autoref{fig:gsm_results}). \textbf{Right:} Matches the baseline's $1024$-step
performance with only a fraction of the sampling steps (see \autoref{tab:best_mauve_steps}, \autoref{fig:various} and \autoref{fig:ap_mauve_ppl_steps}). Sources for each value are listed in \autoref{tab:fig1_sources}.}  
\vspace{-10pt}
  \label{fig:teaser}
\end{figure*}

\section{Introduction}
\label{intro}

\vspace{-5 pt}
Current leading language models~\citep{achiam2023gpt, anil2023gemini, guo2025deepseek} primarily rely on autoregressive decoding, generating one token at a time. While training benefits from the parallelism of transformer architectures and teacher forcing, inference remains fundamentally sequential, creating a major bottleneck as models and sequence lengths scale. Diffusion language models~\citep{khanna2025mercury, nie2025large, sahoo2024simple} have recently emerged as a promising alternative, offering significantly more flexibility in their generative trajectories. By learning to reverse a corruption process applied to the entire sequences simultaneously, they can generate multiple tokens in parallel and potentially enable substantially faster inference.

While most diffusion language models rely on discrete corruption processes such as masking~\citep{sahoo2024simple, arriola2025block} or token replacement~\citep{schiff2024simple, lou2023discrete}, recent work has revisited continuous diffusion and flow-based formulations for language~\citep{dieleman2022continuous, lee2026flow}. Such formulations have been central to advances in image and video generation and provide smooth generative trajectories that support efficient ODE- and SDE-based sampling~\citep{lipman2023flow, song2021score}. They are also particularly well-suited to few-step generation, as continuous trajectories can be effectively approximated with a small number of integration steps~\citep{lee2026flow, kang2025parallelbench}. Importantly, they make techniques developed for continuous generative models directly applicable to language, including guidance~\citep{ho2022classifier, karras2024guiding} and flow-based distillation~\citep{pmlr-v202-song23a, boffi2025flow}. These properties make continuous diffusion a promising and versatile framework for text generation.

Interestingly, recent advances in diffusion modeling for image synthesis have shown that augmenting the standard denoising objective with auxiliary self-supervised signals can substantially improve both training efficiency and generation quality \citep{yu2024repa, kouzelis2025redi, chefer2025selfflow, jiang2025sra}. These methods often involve distilling semantic representations from pretrained encoders \citep{yu2024repa, kouzelis2025redi, wu2025reg}, but this introduces a dependency on external models whose suitability can be difficult to anticipate, especially in the language domain. More recent methods instead exploit self-supervision within the generative model itself, leveraging the structure of the denoising trajectory to define objectives such as contrastive learning or self-distillation across noise levels and network layers \citep{chefer2025selfflow, jiang2025sra, haghighi2025layersync}. These results suggest that the denoising process contains rich internal supervision signals that go beyond direct reconstruction of clean data.

This raises a natural question: \emph{can such self-supervised objectives similarly improve continuous diffusion models for language?}

In this work, we answer this question affirmatively and introduce \texttt{SERF} (\textbf{Se}\textit{lf}-\textbf{R}\textit{egularized} \textbf{F}\textit{low Language Models}), a simple and effective training framework that integrates self-supervised learning into continuous flow-matching language models (see \autoref{fig:method}). We build on Flow Matching Language Models (FLMs)~\citep{lee2026flow}, which operate over one-hot token embeddings and are trained to predict the clean one-hot embedding from a noised input using a cross-entropy loss.
\texttt{SERF} augments this objective with a noisy-to-clean self-distillation signal derived from the flow trajectory. Given an input sequence, we construct two corrupted versions using the same underlying noise but at different noise levels. The student processes the more corrupted input, while an exponential moving average (EMA) teacher processes a cleaner version along the same interpolation path. Since the teacher solves an easier denoising problem, its representations and predictions provide informative targets for what the student should recover from the noisier input. Matching these targets encourages consistency along the denoising trajectory and provides richer supervision, without modifying the model architecture or inference procedure.

Inspired by self-distillation approaches in image diffusion~\citep{chefer2025selfflow, jiang2025sra, haghighi2025layersync}, we study two instantiations of \texttt{SERF}. Following these prior works, \emph{Latent Distillation} (\texttt{SERF-L}) aligns intermediate hidden representations, matching the student's features at a shallow block to the teacher's features at a deeper block through a lightweight projection head. This encourages the student's early layers to form the contextual representations that the teacher reaches deeper in the network and from a cleaner input. 
Our second variant, \emph{Probability Distribution Distillation} (\texttt{SERF-P}), distills the teacher's predicted token distribution instead of its features. This target is natural for language, where the denoiser outputs a categorical distribution over the vocabulary at every position. Beyond matching the student's final output, we also investigate distillation at an early student block: its hidden states pass through a small adapter and the teacher's frozen output head, and are trained with a cross-entropy loss to match the teacher's prediction from the cleaner input. Reusing the teacher's head keeps both distributions in the same output space and encourages the student to recover, already in its early layers, which tokens are plausible given a less corrupted context. This distribution-level target is also closer to the language modeling objective than feature alignment, as it shares the cross-entropy form of the denoising loss.
\texttt{SERF} is fully self-contained: all proposed variants derive their supervision from the flow model itself and require no external teacher or pretrained representation model. Moreover, \texttt{SERF} is used only during training and leaves the underlying sampling procedure unchanged, introducing no additional inference-time cost.

To the best of our knowledge, \texttt{SERF} is the first framework to systematically integrate self-supervised objectives into continuous diffusion language models. Our results demonstrate that simple self-distillation along the denoising trajectory can substantially improve training efficiency, generation quality, and few-step generation.

\paragraph{Contributions.}
Our contributions can be summarized as follows:
\begin{enumerate}[leftmargin=20pt, topsep=0pt, parsep=0pt]
\item We introduce \texttt{SERF}, a simple self-regularization framework that exploits the denoising trajectory of continuous flow language models through noisy-to-clean self-distillation, requiring neither external pretrained models nor changes to the architecture or inference procedure.
\item We investigate two instantiations of \texttt{SERF}: \texttt{SERF-L},
which aligns hidden representations, and \texttt{SERF-P}, which aligns output
token distributions, including at early student layers.

\item We show that \texttt{SERF} generalizes across three continuous diffusion
paradigms (\texttt{FLM}, \texttt{ELF}, and \texttt{S-FLM}), yielding
substantial gains in training time, generation quality, and sampling
efficiency. On \texttt{OpenWebText}, \texttt{SERF} reaches the MAUVE of
\texttt{FLM} with up to $7.7\times$ less wall-clock training time and
ultimately raises MAUVE by $24.3\%$. On \texttt{GSM8K}, \texttt{SERF}  improves the
accuracy of \texttt{S-FLM} by $28.8\%$ and matches its $1024$-step accuracy
with $64\times$ fewer sampling steps (\autoref{fig:teaser}).

\end{enumerate}

\section{Related Work}

\paragraph{Discrete Diffusion for Language.} Early approaches to diffusion-based language modeling operate directly in the discrete token space, defining forward processes based on masking~\citep{sahoo2024simple, arriola2025block, nie2025large, chao2025beyond}  or random token substitution~\citep{schiff2024simple, lou2023discrete, sahoo2025diffusion} and training models to reverse these corruptions. While theoretically principled, discrete diffusion models face fundamental challenges: tractable inference requires factorized approximations of the reverse process, and the discrete nature of the state space limits the applicability of continuous optimization tools~\citep{lee2026flow}. Recent efforts have sought to address these limitations by combining discrete and continuous corruption~\citep{zhou2025coevolutionary, zheng2025continuously, pynadath2025candi}. In contrast, we operate entirely in a continuous space, avoiding discrete approximations at both training and inference time.

\paragraph{Continuous Diffusion for Language.} Continuous diffusion and flow matching language models lift tokens into a continuous representation before applying the generative process, enabling the use of standard continuous optimization. The choice of representation varies across methods: learned token embeddings~\citep{li2022diffusion, gulrajani2023likelihood, gong2022diffuseq, chen2026langflow}, frozen pretrained embeddings~\citep{lovelace2023latent, strudel2022self}, and one-hot vectors~\citep{lee2026flow} have all been explored, alongside Riemannian manifold formulations that respect the geometry of the probability simplex~\citep{jo2025continuous, cheng2024categorical, davis2024fisher}.
Our work builds on three continuous flow-based language models. \texttt{FLM}~\citep{lee2026flow}
operates in the unconstrained Euclidean setting, representing tokens as
one-hot vectors and transporting Gaussian noise to these targets via a
learned velocity field trained with a cross-entropy objective.
\texttt{S-FLM}~\citep{deschenaux2026language} instead defines the flow on
the unit hypersphere, with token embeddings and hidden states constrained
to $\mathbb{S}^{d-1}$. \texttt{ELF}~\citep{hu2026elf} performs flow matching
in a continuous latent space of pretrained \texttt{T5} sentence embeddings.
We show that \texttt{SERF} applies to all three.
A related line of work has explored few-step sampling via flow map distillation~\citep{lee2026flow, Roos2026CategoricalFM}, where a teacher model is used to
distill the generative trajectory into fewer sampling steps. While this  direction also involves a teacher-student setup, it is orthogonal to ours:
our primary goal is not to reduce the number of sampling steps but to improve the
training objective itself through self-regularization, leaving the inference
procedure entirely unchanged.

\paragraph{Self-Supervised Learning in Image Diffusion.}

In image generation, a growing body of work has explored augmenting diffusion and flow model
training with self-supervised signals to improve representation quality and
generation performance. One prominent direction leverages pretrained
self-supervised encoders \citep{radford2021clip, oquab2024dinov2, he2020mocov3, assran2023ijepa} to align intermediate
diffusion features with rich external visual representations, yielding
substantial gains in training efficiency and sample quality~\citep{yu2024repa, kouzelis2025redi, kouzelis2026coevolving, wu2025reg, singh2025irepa, leng2025repae}. A
complementary set of approaches instead derives supervision entirely from within
the generative model itself, through objectives such as  contrastive approaches \citep{stoica2025contrastive, wang2025diffuse}, or teacher-student
alignment along the denoising trajectory \citep{chefer2025selfflow, jiang2025sra, haghighi2025layersync}, without relying on an
external encoder. Although methods based on external representations have
generally achieved strong results, \cite{chefer2025selfflow} showed that purely
self-supervised objectives, without external representations, can exceed their performance and scale more favorably across modalities and model
sizes.
Motivated by these findings, we ask whether auxiliary self-supervised objectives can improve continuous flow matching language models.

\section{Method}
\label{method}

\begin{figure}[h]

    \centering
\includegraphics[width=\textwidth]{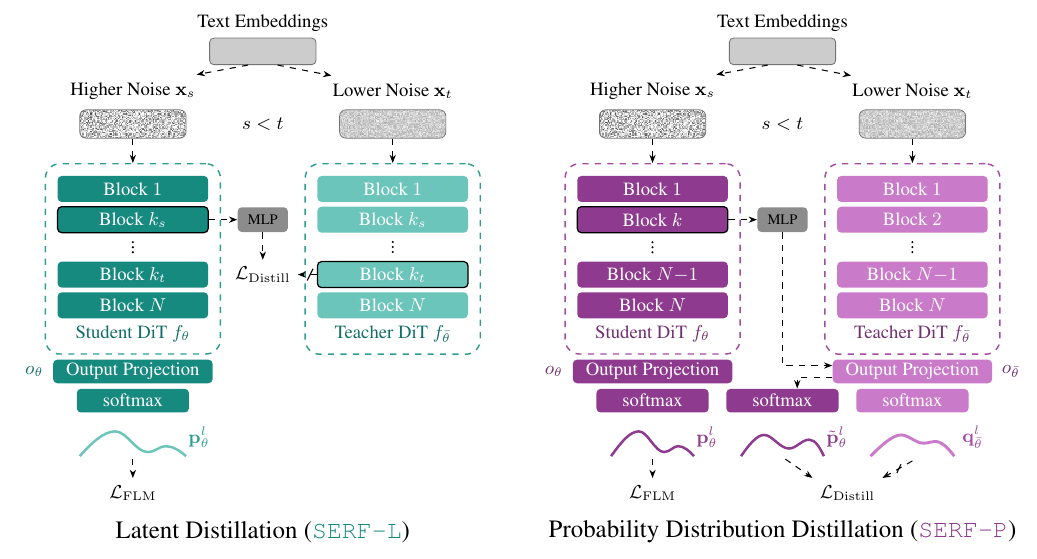}
 \vspace{-20pt}
\caption{\textbf{Overview of \texttt{SERF}.}
In flow language models, the clean data lies at $\mathbf{x}_1$ (one-hot
token embeddings) and pure noise at $\mathbf{x}_0$. \texttt{SERF} corrupts
$\mathbf{x}_1$ at two noise levels along the same interpolation path using
shared noise: the noisier input $\mathbf{x}_s$ is fed to the student
$f_\theta$, while the cleaner input $\mathbf{x}_t$ ($t > s$) is fed to the
EMA teacher $f_{\bar{\theta}}$, whose outputs are detached via
stop-gradient. The student is supervised by the standard cross-entropy loss
against the ground-truth one-hot target $\mathbf{x}_1^l$, plus a
distillation loss weighted by $\lambda$, which takes one of two forms.
\textbf{Left:} In Latent Distillation, (\texttt{SERF-L}) the hidden states of an
early student block $k_s$ are mapped through a lightweight projection head
$g_\theta$ and aligned with those of a deeper teacher block $k_t > k_s$
using a per-token cosine distance.
\textbf{Right:} In Probability Distribution Distillation (\texttt{SERF-P}), features
from student block $k$ pass through a small MLP and the teacher's frozen head
$o_{\bar{\theta}}$, giving token-wise distributions $\tilde{\mathbf{p}}_\theta^l$
that are trained to match the teacher's $\mathbf{q}_{\bar{\theta}}^l$ via a
cross-entropy loss; the student's own output $\mathbf{p}_\theta^l$ is trained with
$\mathcal{L}_\mathrm{FLM}$.
}
    \label{fig:method}
    \vspace{-10pt}
\end{figure}

In this section, we introduce Self-Regularized Flow Language Models (\texttt{SERF}), a training framework that regularizes a flow language model by distilling signals from an EMA teacher evaluated at a cleaner point along the same interpolation path (see \autoref{fig:method}). We first review continuous one-hot flow matching for language modeling (\autoref{sec:prelim}). We then introduce \texttt{SERF} and its two variants, based on latent and probability distribution distillation (\autoref{sec:serf}), before describing the strategies used to select the student and teacher noise levels (\autoref{sec:noise_lvl}). Although we instantiate \texttt{SERF} using \texttt{FLM}~\citep{lee2026flow}, our framework can in principle be applied to other continuous diffusion or flow-based language models.

\subsection{Preliminary: Flow Language Models}
\label{sec:prelim}

Flow matching language models operate in continuous representations; one natural choice, as in \cite{lee2026flow}, is to embed discrete tokens as one-hot vectors. Formally, let $\mathcal{V}$ be a vocabulary and let $\mathbf{y}=(y^1,\ldots,y^L)\in\mathcal{V}^L$ denote a sequence of length $L$. A flow matching language model represents $\mathbf{y}$ as the continuous one-hot matrix
\begin{equation}
    \mathbf{x}_1 = \mathrm{onehot}(\mathbf{y}) \in \mathbb{R}^{L \times |\mathcal{V}|}.
\end{equation}
\noindent \textbf{Training.} Once the tokens have been represented as continuous vectors, a time-dependent interpolation process transforms the data $\mathbf{x}_1 \sim p(x)$ into Gaussian noise $\boldsymbol{\epsilon} \sim \mathcal{N}(0, 1)$:
\begin{equation}
    \mathbf{x}_s = s \mathbf{x}_1 + (1-s)\boldsymbol{e}
\end{equation}
where $s \in[0,1]$.
The denoiser $f_\theta(\mathbf{x}_s,s)$ is parameterized
with a token-wise softmax output layer, ensuring that the predicted
distribution at each position lies on the probability simplex $\Delta^{|\mathcal{V}|-1}$, and outputs
a per-position probability distribution over the vocabulary, i.e.,
$\mathbf{p}_\theta^l = f_\theta(\mathbf{x}_s,s)^l \in \Delta^{|\mathcal{V}|-1}$
for each position $l$. Since $\mathbf{x}_1$ is one-hot, the denoiser is
trained with token-level cross-entropy:
\begin{equation}
    \mathcal{L}_{\mathrm{FLM}}
    =
     \mathbb{E}_{\mathbf{x}_1, \boldsymbol{e}, s}
    \Bigg [
    - \frac{1}{L} \sum_{l=1}^L
    (\mathbf{x}_1^l)^{\mathsf{T}} \cdot \log \mathbf{p}_\theta^l
    \Bigg ].
\end{equation}

\noindent \textbf{Sampling.} During sampling the denoiser is converted into a velocity field by $v_\theta(\mathbf{x}_s,s) = \bigl(f_\theta(\mathbf{x}_s,s)-\mathbf{x}_s\bigr)/(1-s)$, and sampling is performed by integrating the probability flow ODE $ \dot{\mathbf{x}}_s = v_\theta(\mathbf{x}_s,s)$
from $s=0$ to $s=1$, starting from $\mathbf{x}_0\sim\mathcal{N}(\mathbf{0},\mathbf{I})$. The final continuous state $\mathbf{\hat{x}}_1^l$ is decoded back to discrete tokens with a simple argmax operation: $\hat{\mathbf{y}}^l = \arg\max_{v\in\mathcal{V}}\, \mathbf{\hat{x}}_1^l(v)$.

\subsection{Self-Regularized Flow Language Models} \label{sec:serf}

\paragraph{Teacher--student framework.}
Standard \texttt{FLM} training supervises the denoiser only through
cross-entropy against the clean sequence. We add a self-supervised signal
from a teacher derived from the model itself, requiring no external network
and adding no inference cost. Let $f_\theta$ be the student denoiser and
$f_{\bar{\theta}}$ the teacher, whose parameters are an exponential moving
average (EMA) of the student's,
$\bar{\theta} \leftarrow \alpha \bar{\theta} + (1-\alpha)\theta$ with
$\alpha \in (0,1)$. For each sequence $\mathbf{x}_1$, we construct a noisy
student input and a cleaner teacher input at times $s \leq t$:
\begin{equation}
\label{eq:interpolants}
    \mathbf{x}_s = s\,\mathbf{x}_1 + (1-s)\,\boldsymbol{e},
    \qquad
    \mathbf{x}_t = t\,\mathbf{x}_1 + (1-t)\,\boldsymbol{e},
\end{equation}
where the noise $\boldsymbol{e} \sim \mathcal{N}(0, I)$ is shared, so the
two views differ only in noise level.

Since the teacher receives a cleaner input and therefore solves an easier denoising problem, its hidden representations and output predictions provide informative targets for what the student should recover from the noisier input. We train the student with the general objective
\begin{equation}
\label{eq:serf}
\mathcal{L}_{\mathrm{SERF}}  =  \mathcal{L}_{\mathrm{FLM}}  + \lambda\mathcal{L}_{\mathrm{Distill}},
\end{equation}
where $\lambda>0$ controls the strength of the self-distillation signal. We consider two choices for $\mathcal{L}_{\mathrm{Distil}}$: \emph{Latent Distillation} (\texttt{SERF-L}), which aligns intermediate hidden representations, and \emph{Probability Distribution Distillation} (\texttt{SERF-P}), which aligns output token distributions. In both cases, gradients are propagated only through the student, while the teacher is updated exclusively through EMA. Sampling remains identical to \texttt{FLM}.
The full training procedure for all variants is summarized in \autoref{algo:serf_latent}, \autoref{algo_serf_prob} and \autoref{algo:serf_prob_layer}, where highlights indicate the minimal modifications required on top of the baseline \texttt{FLM}.

\subsubsection{Latent Distillation}
\label{sec:serf_latent}

Intermediate representations of a denoiser encode contextual information well before it is resolved into token predictions. We therefore align the student's hidden states with those the teacher computes from the cleaner input. Let
$\mathbf{h}_\theta^{(k_s)}(\mathbf{x}_s, s) \in \mathbb{R}^{L \times d}$
denote the hidden states after the $k_s$-th transformer block of the student,
and $\mathbf{h}_{\bar{\theta}}^{(k_t)}(\mathbf{x}_t, t) \in \mathbb{R}^{L \times d}$
those after the $k_t$-th block of the teacher. Since the two representations
are taken at different depths and noise levels, the student features are
mapped through a lightweight projection head $g_\theta$ before being
compared, while the teacher features are used as is:
\begin{equation}
    \mathbf{z}_\theta^l = g_\theta\big(\mathbf{h}_\theta^{(k_s)}(\mathbf{x}_s, s)^l\big),
    \qquad
    \bar{\mathbf{z}}^l = \operatorname{sg}\big(\mathbf{h}_{\bar{\theta}}^{(k_t)}(\mathbf{x}_t, t)^l\big),
\end{equation}
where $\operatorname{sg}(\cdot)$ denotes stop-gradient.
The latent distillation loss is
\begin{equation}
\label{eq:serf_latent}
\mathcal{L}_{\mathrm{Distill}}^{\mathrm{L}}  =  \mathbb{E}_{\mathbf{x}_1,\boldsymbol{\epsilon},s,t}
\left[
\frac{1}{L}
\sum_{l=1}^{L}
\operatorname{dist}
\left(
\mathbf{z}_\theta^l,
\bar{\mathbf{z}}^l
\right)
\right],
\end{equation}
where $\operatorname{dist}(\cdot,\cdot)$  denotes the distance between feature vectors, for which we use cosine distance in our experiments. \texttt{SERF-L} is obtained by setting  $\mathcal{L}_{\mathrm{Distil}} = \mathcal{L}_{\mathrm{Distil}}^{\mathrm{L}}$ in \autoref{eq:serf}.

Choosing a shallow student block and a deeper teacher block ($k_s < k_t$)
encourages the early layers of the student to already form the contextual
representations that the teacher only reaches deeper in the network and
from a cleaner input. The projection head is used only during training and
is discarded at inference.

\begin{figure}[t]
    \centering
    \includegraphics[width=0.9\textwidth]{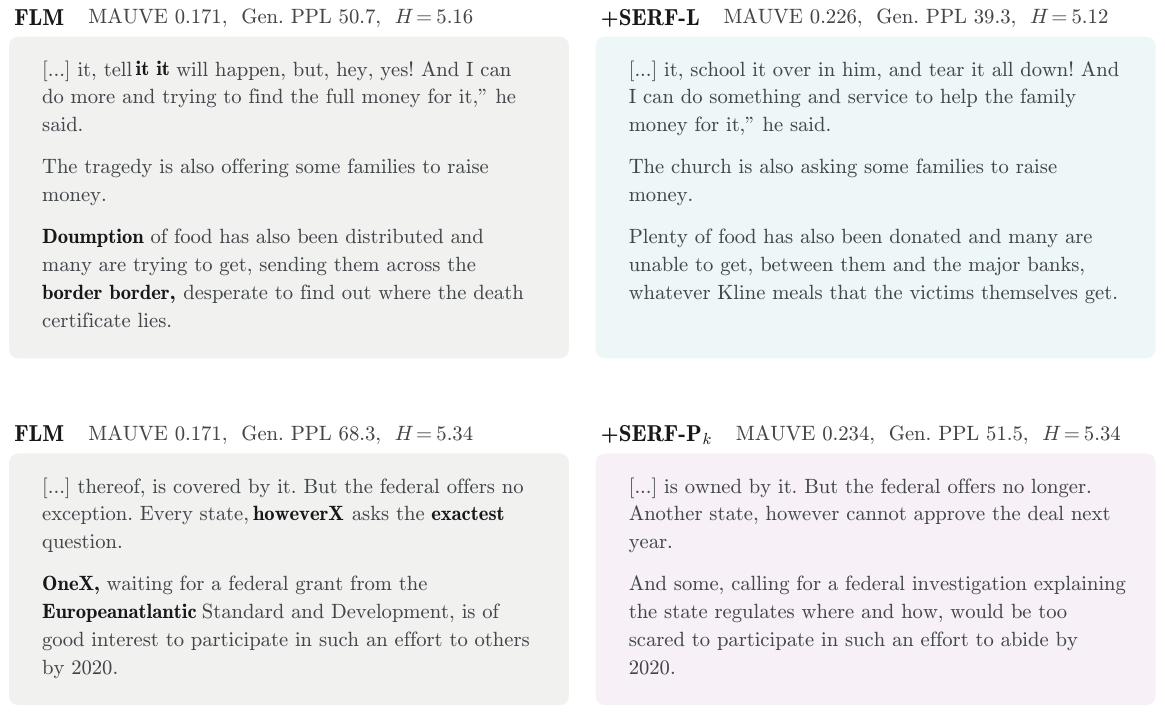}
\caption{\textbf{Generated samples on \texttt{OWT}.} Samples from
\texttt{FLM}, and \texttt{SERF-L}, \texttt{SERF-P}$_k$, trained
for $1$M and $300$K steps and generated with $1024$ sampling steps from the same
initial noise. Errors are highlighted in \textbf{bold}. Both \texttt{SERF}
variants produce notably more fluent text, whereas \texttt{FLM} exhibits
grammatical errors and repetitions. We provide more examples of generated text in \autoref{sec:app_qual}.}

    \label{fig:example_text}
\end{figure}

\subsubsection{Probability Distribution Distillation}
\label{sec:serf_prob}

Beyond internal features, the teacher's predictive distribution itself is an
informative target, since it reflects which tokens remain plausible given a less
corrupted context. We therefore align the student's and the teacher's per-position
distributions, $\mathbf{p}_\theta^l$ and
$\mathbf{q}_{\bar{\theta}}^l
= \operatorname{sg}\!\left(f_{\bar{\theta}}(\mathbf{x}_t, t)^l\right)
\in \Delta^{|\mathcal{V}|-1}$,
via a cross-entropy distillation loss:
\begin{equation}
\label{eq:serf_prob}
\mathcal{L}_{\mathrm{Distill}}^{\mathrm{P}}
=
\mathbb{E}_{\mathbf{x}_1,\boldsymbol{\epsilon},s,t}
\left[
-\frac{1}{L}
\sum_{l=1}^{L}
(\mathbf{q}_{\bar{\theta}}^l)^{\mathsf{T}}
\log\mathbf{p}_\theta^l
\right].
\end{equation}
\texttt{SERF-P} is obtained by setting
$\mathcal{L}_{\mathrm{Distil}} = \mathcal{L}_{\mathrm{Distil}}^{\mathrm{P}}$
in \autoref{eq:serf}, with the student's own output as the aligned distribution,
$\mathbf{p}_\theta^l = f_\theta(\mathbf{x}_s, s)^l$.

\paragraph{Distilling into an early layer.}
\texttt{SERF-L} benefits most from aligning a shallow student block with a
deeper teacher block (see \autoref{fig:ablation_layers}), consistent with findings in
image diffusion~\citep{chefer2025selfflow, jiang2025sra}: supervising early layers
directly encourages the student to form, early in the network, information that
it would otherwise only build at full depth. When the distillation loss acts on
the final output alone, early layers receive this signal only indirectly,
through the entire network. We therefore also apply \autoref{eq:serf_prob} to
an intermediate layer of the student. Let $\mathbf{h}_\theta^{(k)}$ be the hidden states
after the $k$-th block. We map them to a distribution with a lightweight
projection head $g_\theta$ followed by the \emph{teacher's} output head
$o_{\bar{\theta}}$, kept frozen:
\begin{equation}
\label{eq:serf_prob_layer}
\mathbf{\tilde{p}}_\theta^l
= o_{\bar{\theta}}\!\left(g_\theta\!\left(\mathbf{h}_\theta^{(k)}(\mathbf{x}_s, s)\right)\right)^{l},
\end{equation}
which we denote \texttt{SERF-P}$_k$. Reusing the teacher's head, rather than the student's, ensures that both distributions are obtained through the same mapping to token probabilities, while avoiding interference with the student's output head, which is optimized for final-layer predictions. Unlike \texttt{SERF-P} which distills the final predictions through the student's own output head, \texttt{SERF-P}$_k$ supervises intermediate representations through the teacher's output head and a lightweight adapter.

\noindent \textbf{Remark.} We note that \texttt{SERF} is orthogonal to flow map
distillation methods~\citep{lee2026flow, Roos2026CategoricalFM}, whose primary
goal is to reduce the number of inference steps by distilling the generative
trajectory into a fewer-step student. Such methods also typically require a
pretrained teacher model to provide the
distillation targets. Our objective, instead, is to train a regularizer: it
improves the quality of the learned flow without modifying the inference
procedure, and can in principle be combined with any such distillation
method.

\subsection{Student and Teacher Noise Level Strategies.}
\label{sec:noise_lvl}
A key design choice is how to construct the pair $(s, t)$ with $t > s$. We consider two strategies.

\noindent \textbf{Fixed Offset.} The simplest strategy is to sample a student
timestep $s \sim \mathcal{U}(0,1)$ and set the teacher timestep to
\[
    t = \min\{s+\delta,1\},
\]
where $\delta\in(0,1)$ controls how far the teacher is advanced toward the
clean data endpoint. Thus, the EMA teacher is always evaluated on a cleaner
version of the same sample, with a fixed student--teacher noise gap up to
clipping at $t=1$.

\noindent \textbf{Random Offset.} Alternatively, we sample the student and
teacher timesteps randomly while ensuring that the teacher receives the cleaner
input. Specifically, we draw two times $\mu,\kappa \sim \mathcal{U}(0,1)$ and
set
\[
    s = \min(\mu,\kappa), \qquad t = \max(\mu,\kappa).
\]
The student is evaluated at the noisier timestep $s$, while the EMA teacher is
evaluated at the cleaner timestep $t$. This exposes the distillation loss to a
range of student--teacher noise gaps, rather than imposing a fixed offset.

\definecolor{plum}{HTML}{8E3B8F}

\begin{table}[t]
\centering
\small
\newcommand{\pmstd}[1]{\,{\scriptsize\color{black!50}$\pm#1$}}
\setlength{\tabcolsep}{5pt}
\renewcommand{\arraystretch}{1.15}
\caption{\textbf{Best MAUVE} ($\uparrow$, top) on \texttt{OWT} over
$\tau \in \{0.95, 1.0, 1.05\}$, where $\pm$ is the std over 5 MAUVE seeds, and
\textbf{Gen.~PPL} ($\downarrow$, bottom) at the unigram entropy of
\texttt{OWT} text ($H = 5.44$), interpolated over each model's temperature
sweep. \texttt{SERF-L} and \texttt{SERF-P}$_k$, trained
for 300K steps, are compared against \texttt{FLM} trained for 1M steps. Best
in \textbf{bold}, second best \underline{underlined}. We further present the full MAUVE and Gen.~PPL fronts in \autoref{fig:ap_mauve_ppl_steps}.}
\label{tab:best_mauve_steps}
\begin{tabular}{llccccc}
\toprule
 & & \multicolumn{5}{c}{\Th{Sampling steps}} \\
\cmidrule(lr){3-7}
\Th{Model} & \#\Th{Iters} & $64$ & $128$ & $256$ & $512$ & $1024$ \\
\midrule
\rowcolor{gray!10}  \multicolumn{7}{@{}l}{\textit{Best MAUVE} ($\uparrow$)} \\
\addlinespace[1pt]
\texttt{FLM} & $1$M   & $0.260$\pmstd{0.011} & $0.329$\pmstd{0.022} & $0.349$\pmstd{0.014} & $0.345$\pmstd{0.005} & $0.313$\pmstd{0.011} \\
\rowcolor{plum!10} \texttt{SERF-P}$_k$ & $300$K & $\mathbf{0.341}$\pmstd{0.019} & $\underline{0.414}$\pmstd{0.026} & $\underline{0.399}$\pmstd{0.010} & $\underline{0.409}$\pmstd{0.026} & $\underline{0.388}$\pmstd{0.004} \\
\rowcolor{teal!10} \texttt{SERF-L} & $300$K & $\underline{0.337}$\pmstd{0.009} & $\mathbf{0.435}$\pmstd{0.015} & $\mathbf{0.418}$\pmstd{0.017} & $\mathbf{0.423}$\pmstd{0.019} & $\mathbf{0.389}$\pmstd{0.005} \\
\midrule
\rowcolor{gray!10} \multicolumn{7}{@{}l}{\textit{Gen.~PPL at $H = 5.44$} ($\downarrow$)} \\
\addlinespace[1pt]
\texttt{FLM} & $1$M   & $86.2$ & $78.8$ & $75.7$ & $74.7$ & $74.9$ \\
\rowcolor{plum!10} \texttt{SERF-P}$_k$ & $300$K & $\underline{83.9}$ & $\underline{76.2}$ & $\underline{73.2}$ & $\underline{72.5}$ & $\underline{73.5}$ \\
\rowcolor{teal!10} \texttt{SERF-L} & $300$K & $\mathbf{82.3}$ & $\mathbf{75.4}$ & $\mathbf{72.6}$ & $\mathbf{71.7}$ & $\mathbf{72.9}$ \\
\bottomrule
\end{tabular}
\end{table}

\section{Experiments}
\label{sec:exp}
To validate \texttt{SERF}, we evaluate its two variants, \texttt{SERF-L}
(Latent Distillation, \autoref{sec:serf_latent}) and \texttt{SERF-P}
(Probability Distribution Distillation, \autoref{sec:serf_prob}), on
unconditional language modeling and math reasoning. We apply them to three
baseline models: \texttt{FLM}~\citep{lee2026flow},
\texttt{S-FLM}~\citep{deschenaux2026language} and
\texttt{ELF}~\citep{hu2026elf}. These span three distinct continuous
paradigms, namely one-hot embeddings, spherical embeddings, and latent
embeddings, underscoring the broad applicability of our method and showing
that it is not tied to the choice of input representation. We first
describe the experimental setup (\autoref{sec:exp_setup}). We then present
unconditional text generation results on \texttt{OWT} with \texttt{FLM} and
\texttt{ELF} (\autoref{sec:exp_uncond}), followed by math reasoning results on
\texttt{GSM8K}\citep{cobbe2021gsm8k} with \texttt{S-FLM} (\autoref{sec:exp_gsm8k}). Finally, we
analyze the effect of the noise level strategy and the choice of distilled layers in (\autoref{sec:exp_analysis}).

\subsection{Experimental Setup}
\label{sec:exp_setup}

\noindent \textbf{Training.}
All training hyperparameters are kept identical to those of the corresponding
baselines. Both \texttt{SERF-L} and \texttt{SERF-P} use the \texttt{random}
offset strategy, with regularization weights $\lambda = 2.0$ and
$\lambda = 0.1$, respectively. For \texttt{SERF-L}, the alignment loss is
active from the start of training, and unless otherwise noted, the student's
block $k_s = 2$ is aligned with the teacher's block $k_t = 10$.
 For \texttt{SERF-P}$_k$, the student's
block is set to $k = 2$.
For \texttt{SERF-P}, we found it is better to initialize the distillation loss after the first $10$K steps. The decay of the EMA teacher is set to $\alpha = 0.9999$.

\noindent \textbf{Evaluation.} We sample 5000 sequences per model and report MAUVE
($\uparrow$)~\citep{pillutla2021mauve} against an equal number of validation
sequences, averaged over 5 seeds, and generative perplexity (Gen.\ PPL
$\downarrow$) under \texttt{GPT-2 Large}~\citep{radford2019language}. We treat
MAUVE as our primary metric: Gen.\ PPL is known to reward repetitive,
degenerate text, and recent work has shown it can be
directly gamed~\citep{franca2026hacking}. To further account for the quality--diversity
trade-off, we compare models not at a single operating point but along
MAUVE--entropy and Gen.\ PPL--entropy fronts, obtained by sweeping the
sampling temperature $\tau$ and measuring average per-sample unigram entropy.
Details on math reasoning evaluation can be found in \autoref{sec:additional_impl}.

\begin{figure}[t]
  \centering
  \includegraphics[width=0.85\textwidth]{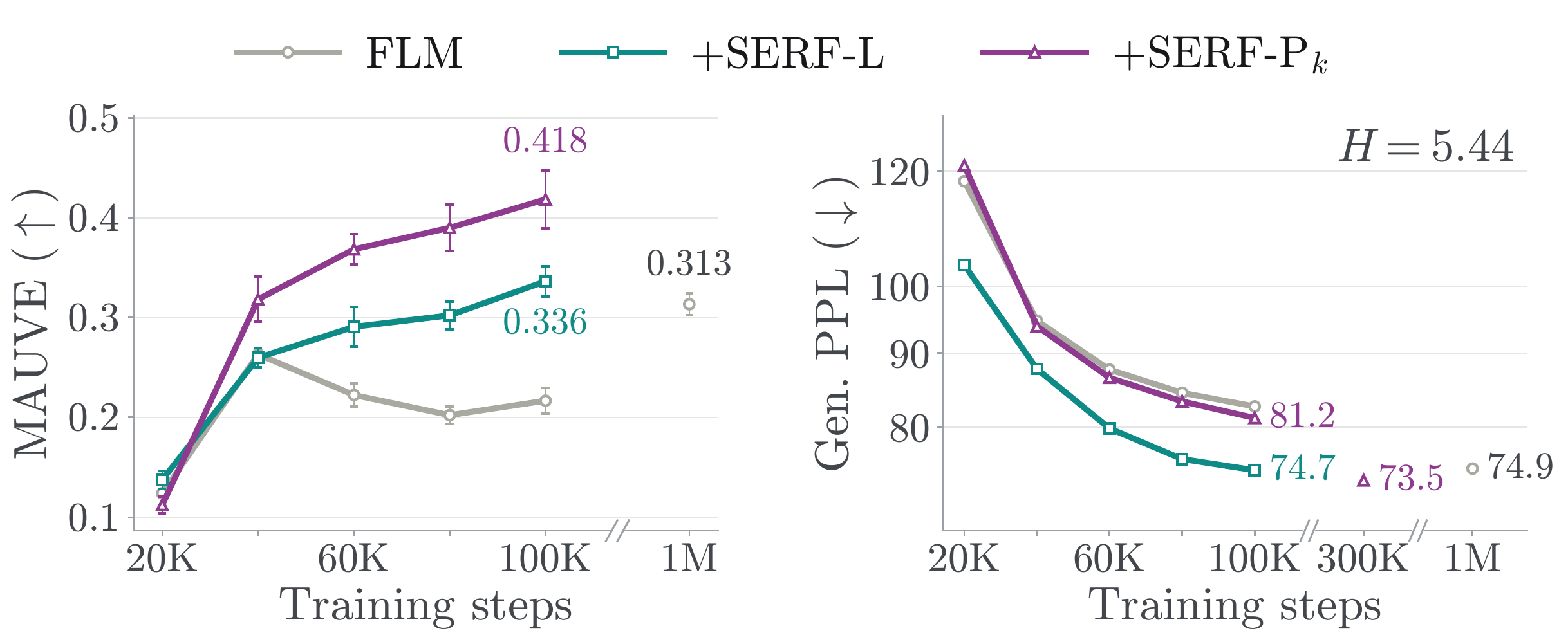}
\caption{\textbf{Training convergence on \texttt{OWT}}.
\texttt{SERF-L} and \texttt{SERF-P} compared against \texttt{FLM}. At each
iteration, we report the best MAUVE over $\tau \in \{0.95, 1.0, 1.05\}$ and
the Gen.~PPL interpolated to the dataset entropy (see \autoref{sec:additiona_eval_details}), using $1024$ sampling steps.
 }
  \label{fig:various}
   \vspace{-15pt}
\end{figure}

\subsection{Unconditional Text Generation Results}
\label{sec:exp_uncond}
To demonstrate the effectiveness of our approach, we evaluate two variants of
\texttt{SERF} on \texttt{OpenWebText} (\texttt{OWT})~\citep{gokaslan2019openwebtext}:
\texttt{SERF-L} (Latent Alignment) and \texttt{SERF-P}, \texttt{SERF-P}$_k$  (Probability
Distribution Alignment). We apply both to the \texttt{FLM} baseline, and
additionally apply \texttt{SERF-L} to \texttt{ELF} \citep{hu2026elf}.

\paragraph{Results with Flow Language Models.}
Both variants substantially outperform \texttt{FLM} across all sampling
budgets from $64$ to $1024$ steps (\autoref{tab:best_mauve_steps}), while
requiring $300$K training steps instead of $1$M. For example, at $1024$ steps
they raise MAUVE from $0.313$ to $0.388$ (\texttt{SERF-P}$_k$) and $0.389$
(\texttt{SERF-L}). The gains in Gen.~PPL point in the same direction: both
variants achieve lower perplexity than \texttt{FLM} at every sampling budget,
with the largest margin in the few-step regime ($86.2 \rightarrow 82.3$ for
\texttt{SERF-L} at $64$ steps). With only $256$ steps, both variants already
reach a lower perplexity ($72.6$ and $73.2$) than \texttt{FLM} at any budget
($74.7$ at best). Since Gen.~PPL is evaluated at a matched unigram entropy,
these gains do not come from reduced sample diversity.

The training curves in \autoref{fig:various} make this explicit: by
$100$K steps, both variants exceed the $1$M-step \texttt{FLM} in MAUVE
($0.336$ for \texttt{SERF-L} and $0.418$ for \texttt{SERF-P}$_k$, vs.\
$0.313$). In Gen.~PPL, \texttt{SERF-L} stays well below \texttt{FLM}
throughout and matches the $1$M-step model at $100$K steps ($74.7$ vs.\
$74.9$), while \texttt{SERF-P}$_k$ tracks \texttt{FLM} early on and surpasses
it at $300$K steps ($73.5$). \autoref{fig:example_text} shows the improvement
qualitatively.
\vspace{-10pt}
\paragraph{Results with Embedded Language Flows.}
To test whether \texttt{SERF} transfers beyond \texttt{FLM}, we apply it to
\texttt{ELF-B}, training both the baseline and our variant for $5$ epochs as in \cite{hu2026elf}. We
evaluate only \texttt{SERF-L}, as \texttt{SERF-P} is not applicable here:
\texttt{ELF} does not output a probability distribution over tokens. \autoref{fig:gsm_results} (right) reports MAUVE against
unigram entropy with $64$ sampling steps, where each curve sweeps the
self-conditioning guidance scale $w$. \texttt{SERF-L} improves over
\texttt{ELF-B} at every guidance scale, shifting the whole front upwards, with
the gains most pronounced at intermediate scales ($w=1$ and $w=1.5$). This
shows that the improvements of latent distillation carry over to a different
continuous diffusion language model paradigm.

\definecolor{plum}{HTML}{8E3B8F}

\begin{figure}[t]
  \centering
  \begin{minipage}[c]{0.25\textwidth}
    \centering
    \footnotesize
    \setlength{\tabcolsep}{3pt}
    \renewcommand{\arraystretch}{0.9}
    \begin{tabular}{@{}lr@{}}
      \toprule
      Model & Acc. (\%) \\
      \midrule
      \multicolumn{2}{@{}l}{\textit{Autoregressive}} \\
      \quad \texttt{Sample} & $53.9$ \\
      \quad \texttt{Greedy} & $\underline{63.3}$ \\
      \midrule
      \multicolumn{2}{@{}l}{\textit{Discrete Diffusion}} \\
      \quad \texttt{MDLM} & $18.0$ \\
      \quad \texttt{DUO}  & $17.2$ \\
      \midrule
      \multicolumn{2}{@{}l}{\textit{Continuous Diffusion}} \\
      \quad \texttt{CANDI} & $0.2$ \\
      \quad \texttt{FLM}   & $0.3$ \\
      \quad \texttt{S-FLM} & $18.0$ \\
      \rowcolor{gray!10}\quad \texttt{S-FLM}$^*$ & $17.7$ \\
      \rowcolor{teal!10} \quad +\texttt{SERF-L} & $22.3$ \\
       \rowcolor{plum!10} \quad +\texttt{SERF-P} & $20.2$ \\
      \rowcolor{plum!10} \quad +\texttt{SERF-P}$_k$ & $\mathbf{22.8}$ \\
      \bottomrule
    \end{tabular}
  \end{minipage}\hfill
  \begin{minipage}[c]{0.74\textwidth}
    \centering
    \includegraphics[width=\linewidth]{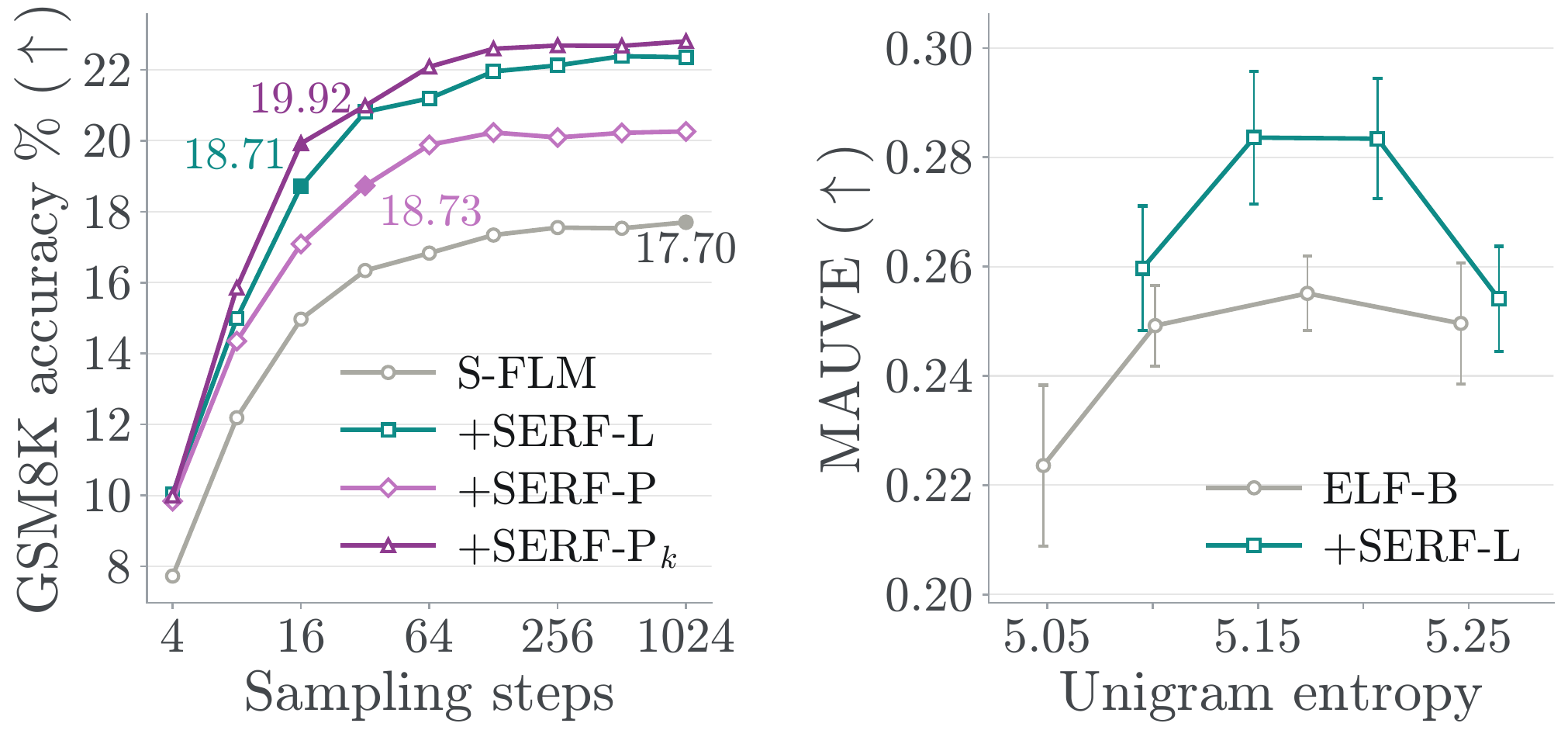}
  \end{minipage}
  \caption{\textbf{Left: GSM8K accuracy} at $250$K training steps. Diffusion models
    at $1024$ steps with $\tau = 1$. Best overall \underline{underlined}, best
    continuous diffusion \textbf{bolded}. $^*$Denotes our evaluation.
    \textbf{Middle:} \texttt{GSM8K} accuracy across sampling steps.
\texttt{SERF-L}, \texttt{SERF-P}, and \texttt{SERF-P}$_k$ compared against \texttt{S-FLM}, all
trained on \texttt{TinyGSM}~\citep{liu2023tinygsm} for $250$K steps. We report the mean over 5 sample seeds.
    \textbf{Right:} Comparison with \texttt{ELF} on \texttt{OWT}.
\texttt{SERF-L} compared against \texttt{ELF-B}, both trained for $5$ epochs
($\sim$$95$K steps) and sampled with $64$ steps, sweeping the
self-conditioning guidance scale $w \in \{0.5, 1.0, 1.5, 2.0\}$. }
  \label{fig:gsm_results}
\end{figure}

\subsection{Reasoning on GSM8K}
\label{sec:exp_gsm8k}

\textbf{Accuracy Improvements.} We apply \texttt{SERF-L}, \texttt{SERF-P} and
\texttt{SERF-P}$_k$ to \texttt{S-FLM}~\citep{deschenaux2026language}, train on
\texttt{TinyGSM}, and evaluate accuracy on the \texttt{GSM8K} test set.
Following \texttt{S-FLM}, we report accuracy at $\tau = 1$ with $1024$
sampling steps for the diffusion models. As shown in
\autoref{fig:gsm_results} (left), all three variants improve the accuracy of
\texttt{S-FLM} from $17.7\%$ to $22.3\%$ (\texttt{SERF-L}), $20.2\%$
(\texttt{SERF-P}) and $22.8\%$ (\texttt{SERF-P}$_k$), with
\texttt{SERF-P}$_k$ achieving the best result among continuous diffusion
language models, while \texttt{CANDI} and \texttt{FLM} remain near zero. All
three variants also outperform discrete diffusion models at $\tau = 1$. A gap
remains under low-temperature decoding ($\tau = 0.1$), where \texttt{MDLM} and
\texttt{DUO} reach $33$--$36\%$~\citep{deschenaux2026language}.

\textbf{Sampling acceleration.} The improvement holds at every sampling budget,
as shown in \autoref{fig:gsm_results} (middle), which substantially reduces the
number of steps needed to reach a given accuracy. \texttt{SERF-P}$_k$ and
\texttt{SERF-L} surpass the $1024$-step accuracy of \texttt{S-FLM} ($17.7\%$)
with only $16$ sampling steps ($19.9\%$ and $18.7\%$, respectively), a
$64\times$ reduction in sampling cost, and \texttt{SERF-P} does so with $32$
steps ($18.7\%$), a $32\times$ reduction.

\newpage
\subsection{Analysis}
\label{sec:exp_analysis}
\begin{figure}[t]
    \centering
    \includegraphics[width=\textwidth]{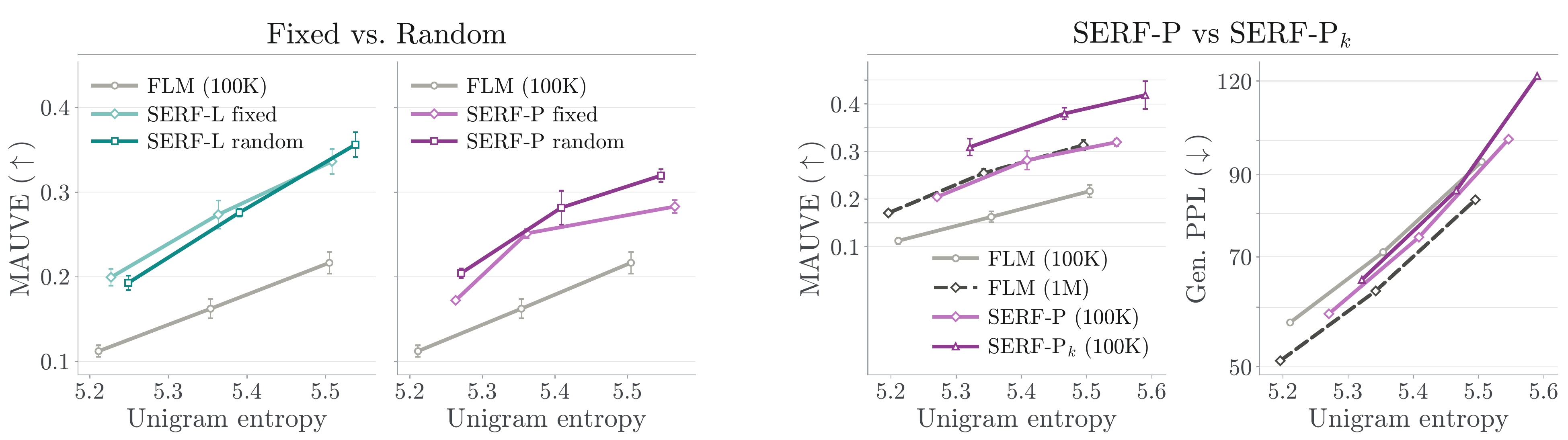}
\caption{ \textbf{Left:}
\texttt{fixed} ($\delta=0.8$) vs.\ \texttt{random} offset strategy for \texttt{SERF-L} and
\texttt{SERF-P}, respectively.
\textbf{Right:} The teacher's output
distribution supervises the student's final layer (\texttt{SERF-P}) or an
earlier block (\texttt{SERF-P}$_k$, $k = 2$). Both trained for $100$K steps
on \texttt{OWT}.
}
    \label{fig:ablation_noise_weight}
\end{figure}

\begin{wrapfigure}{l}{0.3\textwidth}
    \centering
    \vspace{-10pt}
    \includegraphics[width=\linewidth]{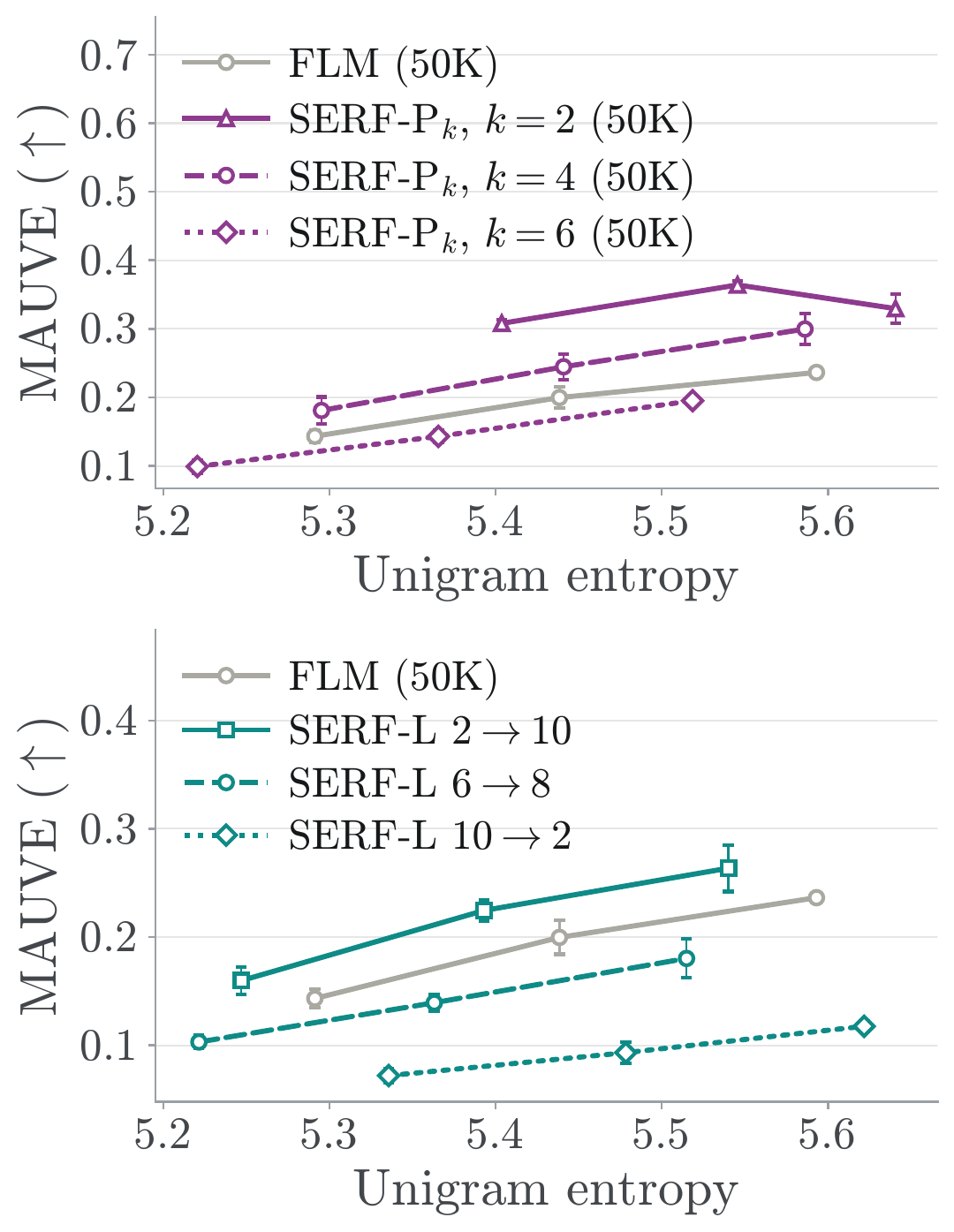}
    \caption{Layer ablations for \texttt{SERF-L} and \texttt{SERF-P}$_k$. Models trained for $50$K.}
    \vspace{-10pt}
    \label{fig:ablation_layers}
\end{wrapfigure}

\paragraph{Effect of Layer choice.}
In \autoref{fig:ablation_layers} (bottom), we study which layers to align in
\texttt{SERF-L}, denoting by $k_s \to k_t$ the alignment of student
layer $k_s$ to teacher layer $k_t$, with all models trained on
\texttt{OWT} for $50$K steps. The choice of layers has a clear effect.
Aligning an early student layer to a late teacher layer ($2\to10$) gives the
highest MAUVE across the whole entropy range, followed by aligning two
intermediate layers ($6\to8$), while the reversed configuration ($10\to2$),
which aligns a late student layer to an early teacher layer, performs worst.
This is consistent with prior work on image generation ~\citep{chefer2025selfflow, jiang2025sra}, which observed
that aligning an early layer of the student to a later layer of the teacher
is beneficial: it encourages the student to form, already in its early
layers, the higher-level representations that the teacher builds deeper in
the network.

 We observe a similar behavior when distilling the teacher's predicted
distribution into an intermediate student layer.
\autoref{fig:ablation_layers} (top) varies the student block $k$ of
\texttt{SERF-P}$_k$, whose hidden state is projected to the vocabulary and
matched to the teacher's prediction ($50$K steps). Earlier blocks are
consistently better: $k=2$ reaches the highest MAUVE ($0.36$ at peak),
$k=4$ is lower but still above \texttt{FLM}, and $k=6$ falls below the
\texttt{FLM} baseline at every entropy. Distilling into early layers thus
helps for both latent and distribution targets, with $k=2$ the best choice.

\paragraph{Distilling distributions into an early layer.}
\autoref{fig:ablation_noise_weight} (right) compares applying the teacher's
distribution to the student's output (\texttt{SERF-P}) with applying it to an
early block (\texttt{SERF-P}$_k$, $k=2$), both trained for $100$K steps on
\texttt{OWT}. \texttt{SERF-P} already matches the MAUVE of \texttt{FLM}
trained for $1$M steps. \texttt{SERF-P}$_k$ improves further: its MAUVE front
lies above all other models across the entire entropy range, and it extends
to higher entropies, producing more diverse samples over the same temperature
sweep. \texttt{SERF-P}$_k$ thus combines the benefits of both variants: like
\texttt{SERF-L}, it supervises early layers directly, but it keeps a
distribution-level cross-entropy loss, of the same form as
$\mathcal{L}_{\mathrm{FLM}}$, rather than aligning hidden features.

\paragraph{Effect of the noise level strategy.}
In \autoref{fig:ablation_noise_weight} (Left) we compare the \texttt{fixed} ($\delta = 0.8$)
and \texttt{random} offset strategies (see \autoref{sec:noise_lvl}) for
\texttt{SERF-L} and \texttt{SERF-P} trained for $100$K steps on \texttt{OWT}.
\texttt{SERF-L} is robust to this choice: both strategies yield nearly
identical fronts that lie well above the $100$K-step \texttt{FLM} baseline
across the full entropy range. \texttt{SERF-P}, in contrast, is sensitive to
it: the \texttt{random} strategy clearly outperforms the baseline, whereas
the \texttt{fixed} offset brings no improvement. We attribute this to the
clipping $t = \min\{s + \delta, 1\}$: with $\delta = 0.8$, the teacher
receives the fully clean input $\mathbf{x}_1$ whenever $s > 0.2$. There, its predictions become nearly one-hot, so
the distillation target of \texttt{SERF-P} adds little beyond the
ground-truth label, whereas \texttt{SERF-L} aligns hidden representations,
which remain informative even for clean inputs.

\section{Conclusion}
In this work, we introduced \texttt{SERF}, a simple self-regularization
framework for continuous diffusion language models. Building on the
observation that an EMA teacher evaluated at a cleaner point on the same
interpolation path provides informative targets for a noisier student,
\texttt{SERF} augments the standard training objective with a noisy-to-clean
self-distillation loss that requires no external models and leaves the
architecture and sampling procedure unchanged. We studied two
instantiations: \texttt{SERF-L}, which distills intermediate hidden
representations, and \texttt{SERF-P}, which distills output token
distributions, together with its early-layer variant \texttt{SERF-P}$_k$.
For both latent and distribution targets, supervising early student layers
proved most effective. Across three baselines, \texttt{SERF} consistently
improves generation quality. On \texttt{OWT}, \texttt{SERF-L} and
\texttt{SERF-P}$_k$ match or exceed \texttt{FLM} trained for $1$M steps using
roughly a tenth of the training budget, with gains that persist in the
few-step regime; on \texttt{GSM8K}, \texttt{SERF} improves the accuracy of
\texttt{S-FLM} and outperforms prior continuous diffusion language models.
Overall, these results suggest that the denoising trajectory is a rich source
of self-supervision for language, and that simple self-distillation along it
is an effective and practical way to exploit it.

\subsection*{AI use statement}
We used large language models (LLMs) as writing assistants during the
preparation of this manuscript, primarily to polish the language, improve
clarity and concision, and help with \LaTeX{} formatting. All scientific
content, including the method, experimental design, results and their
interpretation, was conceived and produced by the authors.

\subsection*{Ethics statement}
Generative models carry a substantial risk of misuse. Their application can lead to various negative societal impacts, most notably the spread of disinformation. Enhancements in generative performance, as achieved by our method, may further increase the realism of generated content, potentially making disinformation even more convincing.

\subsection*{Reproducibility statement}
\texttt{SERF} is fully specified in \autoref{sec:serf} and
Algorithms~\ref{algo:serf_latent} and~\ref{algo_serf_prob}. All
hyperparameters, as well as the architecture, training and sampling details
of each baseline, are reported in \autoref{sec:exp_setup} and the appendix.
All experiments use publicly available datasets, and we will release our
code and checkpoints upon publication.

\bibliography{main}
\bibliographystyle{iclr2027_conference}

\newpage

{\huge \textbf{Appendix}}

\appendix

\section{Additional Details}
\label{sec:additional_impl}

\subsection{Flow Language Models}
\label{app:flm_details}

\paragraph{Architecture.} For all \texttt{FLM}~\citep{lee2026flow}
experiments, we train a 179M-parameter Diffusion Transformer
(\texttt{DiT})~\citep{peebles2023scalable} with 12 transformer blocks, rotary
positional embeddings (\texttt{RoPE})~\citep{su2021roformer}, and adaptive
layer normalization (\texttt{AdaLN}) for time conditioning.

\paragraph{Training.} We train on \texttt{OWT} with a batch size of $512$.
Following \citet{lee2026flow}, we use $2{,}500$ warmup steps and then keep
the learning rate fixed at $3 \times 10^{-4}$, using the Adam
optimizer~\citep{kingma2014adam} with $\beta_1 = 0.9$ and $\beta_2 = 0.999$.
We additionally apply softcapping~\citep{riviere2024gemma2} to the attention
logits, which limits large values and improves numerical stability.

\paragraph{Sampling.} We use an Euler solver followed by a per-position
argmax. At each position $l$, the denoiser outputs a categorical distribution
over the vocabulary,
$\mathbf{p}_\theta^{\,l}(\mathbf{x}_s,s) = \mathrm{softmax}(\boldsymbol{\ell}_\theta(\mathbf{x}_s,s)^l)$,
where $\boldsymbol{\ell}_\theta$ are the network logits, so that $f_\theta$ is
the stack of the $\mathbf{p}_\theta^{\,l}$. We temper these with a temperature
$\tau$,
$\mathbf{p}_\theta^{\,\tau,l} = \mathrm{softmax}(\boldsymbol{\ell}_\theta^{\,l} / \tau)$,
and use the tempered distributions in place of $\mathbf{p}_\theta^{\,l}$ when
forming the velocity field at \emph{every} integration step. Values $\tau < 1$
sharpen the distributions towards the vertices of the simplex and yield
lower-entropy samples, $\tau > 1$ increases diversity, and $\tau = 1$ recovers
the untempered sampler. Sweeping $\tau$ thus traces a quality--diversity
front.

\subsection{Embedded Language Flows}
\label{app:elf_details}

\paragraph{Architecture.} For all \texttt{ELF}~\citep{hu2026elf} experiments,
we use the \texttt{ELF-B} backbone, a 105M-parameter
\texttt{DiT}~\citep{peebles2023scalable} with 12 transformer blocks (hidden
size 768, 12 heads), \texttt{RoPE}~\citep{su2021roformer}, and prefix tokens
for time and self-conditioning. The model operates on
\texttt{T5-small} sentence embeddings.

\paragraph{Training.} We train on \texttt{OWT} with a batch size of $512$.
Following \citet{hu2026elf}, we use $9{,}508$ warmup steps (half an epoch) and
then keep the learning rate fixed at $2 \times 10^{-3}$, obtained by linearly
scaling a base rate of $1 \times 10^{-3}$ with the batch size. Training uses
the Muon optimizer~\citep{jordan2024muon}.

\paragraph{Sampling.} Following \citet{hu2026elf}, we employ an SDE-like
sampler with $64$ steps. Since \texttt{ELF} is trained with a randomly drawn
self-conditioning guidance scale supplied as a conditioning token, the scale
$w$ is set at inference in a single forward pass: small $w$ yields
higher-entropy samples and large $w$ sharpens them, so sweeping
$w \in \{0.5, 1.0, 1.5, 2.0\}$ traces a quality--diversity front.

\subsection{Hyperspherical Flow Language Models}
\label{app:sflm_details}

\paragraph{Architecture.} For all \texttt{S-FLM}~\citep{deschenaux2026language}
experiments, we train a 163M-parameter hyperspherical transformer
(\texttt{S-arch}) with 12 blocks, \texttt{RoPE}~\citep{su2021roformer}, and
time conditioning. Embeddings and hidden states lie on the unit hypersphere
$\mathbb{S}^{d-1}$, and all weight matrices are renormalized after each
optimizer step.

\paragraph{Training.} We train on \texttt{TinyGSM} with a batch size of
$512$. Following \citet{deschenaux2026language}, we use $2{,}500$ warmup steps
and then keep the learning rate fixed at $3 \times 10^{-4}$, using the Adam
optimizer~\citep{kingma2014adam} with $\beta_1 = 0.9$ and $\beta_2 = 0.999$.
Softcapping is not required: stability follows from the normalized
architecture, which unit-normalizes queries and keys and renormalizes all
weight matrices after each optimizer step.

\paragraph{Sampling.} Following \citet{deschenaux2026language}, we use an
Euler solver. The problem statement is fixed to its clean embeddings and
re-projected after every step, so the flow acts only on the answer. At each
position, the denoiser outputs a categorical distribution over the
vocabulary, which determines the direction in which the current embedding
moves; we truncate this distribution at every step to its single
highest-probability token, so the embedding moves directly towards that
token. We decode with an argmax after the final step.

\begin{figure}[t]
    \centering
    \includegraphics[width=\textwidth]{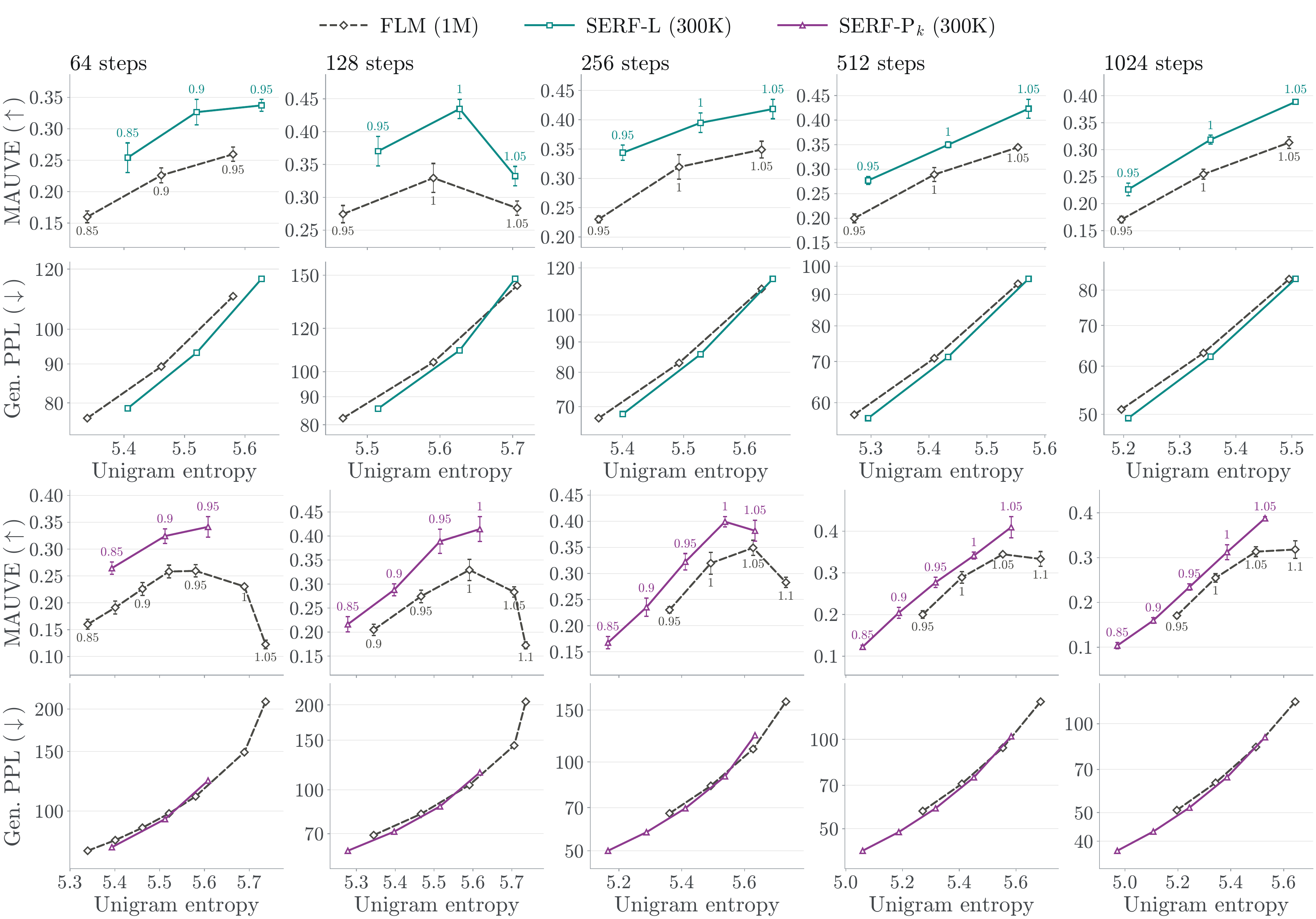}
\caption{\textbf{Mauve and Gen.PPL fronts} for \texttt{SERF-L} and \texttt{SERF-P}$_k$ at 300K steps and for the baseline \texttt{FLM} at 1M steps. Curves obtained by sweeping the sampling temperature $\tau$.}

    \label{fig:ap_mauve_ppl_steps}
\end{figure}

\subsection{Additional Details on SERF} \label{sec:additiona_serf_details}

\paragraph{Implementation Details.}
Unless otherwise noted, the following settings are shared across all three
baselines (\texttt{FLM}, \texttt{ELF} and \texttt{S-FLM}). In both variants
of \texttt{SERF}, the teacher is an EMA of the model weights (decay
$\alpha = 0.9999$), evaluated without gradients on the same noised sequence
as the student but at a cleaner time. The regularization weight is
$\lambda = 2.0$ for \texttt{SERF-L} and $\lambda = 0.1$ for \texttt{SERF-P}.
For \texttt{SERF-L}, the output of student block $k_s = 2$ is passed through a
two-layer MLP projection head  ($768 \to 1536 \to 768$, SiLU) and aligned with the
output of teacher block $k_t = 10$ using a per-token cosine distance. The same MLP architecture is used for \texttt{SERF-P}$_k$. The
projection head is used only during training and discarded at inference.
For \texttt{SERF-P} and \texttt{SERF-P}$_k$, the distillation loss is enabled after an initial
warm-up phase of $10$K training steps. Training algorithms for \texttt{SERF-L} and \texttt{SERF-P}, assuming \texttt{FLM} as the base model, are presented in \autoref{algo:serf_latent} and \autoref{algo_serf_prob} respectively.

\begin{algorithm}
\caption{\texttt{SERF-L} Training (Latent Distillation)}
\label{algo:serf_latent}
\small

\begin{algorithmic}
\Require Dataset $\mathcal{D}$, $\lambda$, $\alpha$, \textcolor{teal}{blocks $k_s$, $k_t$}
\State \textbf{Initialize:} Student and Teacher $f_\theta$, $f_{\bar{\theta}}$, $\bar{\theta} \leftarrow \theta$, \textcolor{teal}{projection head $g_\theta$}
\Repeat
    \State $\mathbf{x}_1 \leftarrow \mathrm{onehot}(\mathbf{y}),\ \mathbf{y} \sim \mathcal{D}$;\quad $\boldsymbol{\epsilon} \sim \mathrm{N}(0, I)$
    \State Sample $s$, $\textcolor{teal}{t}$ \hfill $\triangleright$ See \autoref{sec:noise_lvl}
    \State $\mathbf{x}_s \leftarrow s\mathbf{x}_1 + (1-s)\boldsymbol{\epsilon}, \; \; \textcolor{teal}{\mathbf{x}_t \leftarrow t\mathbf{x}_1 + (1-t)\boldsymbol{\epsilon}}$
    \State $\mathbf{p}_\theta \leftarrow f_\theta(\mathbf{x}_s,s)$
    \State \textcolor{teal}{$\mathbf{z}_\theta \leftarrow g_\theta\big(\mathbf{h}_\theta^{(k_s)}(\mathbf{x}_s,s)\big), \;\; \bar{\mathbf{z}} \leftarrow \operatorname{sg}\big(\mathbf{h}_{\bar{\theta}}^{(k_t)}(\mathbf{x}_t,t)\big)$}
    \State $\mathcal{L}_{\mathrm{FLM}} = - \frac{1}{L} \sum_l (\mathbf{x}_1^l)^\top \log \mathbf{p}_\theta^l$
    \State $\textcolor{teal}{\mathcal{L}_{\mathrm{Distill}} = \frac{1}{L}\sum_l \operatorname{dist}\big(\mathbf{z}_\theta^l, \bar{\mathbf{z}}^l\big)}$
    \State Update $\theta$: $\mathcal{L}_{\mathrm{SERF}} = \mathcal{L}_\mathrm{FLM} + \textcolor{teal}{\lambda \mathcal{L}_\mathrm{Distil}}$
    \State \textcolor{teal}{Update $\bar{\theta} \leftarrow \alpha \bar{\theta} + (1-\alpha) \theta$}
\Until{converged}
\end{algorithmic}
\end{algorithm}

\begin{algorithm}
\caption{\texttt{SERF-P} Training (Probability Distribution Distillation)}
\label{algo_serf_prob}
\small

\begin{algorithmic}
\Require Dataset $\mathcal{D}$, $\lambda$, $\alpha$
\State \textbf{Initialize:} Student and Teacher $f_\theta$, $f_{\bar{\theta}}$, $\bar{\theta} \leftarrow \theta$,
\Repeat
    \State $\mathbf{x}_1 \leftarrow \mathrm{onehot}(\mathbf{y}),\ \mathbf{y} \sim \mathcal{D}$;\quad $\boldsymbol{\epsilon} \sim \mathrm{N}(0, I)$
    \State Sample $s$, $\textcolor{plum}{t}$ \hfill $\triangleright$ See \autoref{sec:noise_lvl}

    \State  $\mathbf{x}_s \leftarrow s\mathbf{x}_1 + (1-s)\boldsymbol{\epsilon}, \; \; \textcolor{plum}{\mathbf{x}_t \leftarrow t\mathbf{x}_1 + (1-t)\boldsymbol{\epsilon}}$
    \State  $\mathbf{p}_\theta \leftarrow f_\theta(\mathbf{x}_s,s), \;\; \textcolor{plum}{\mathbf{q}_{\bar{\theta}} \leftarrow \operatorname{sg} \big( f_{\bar{\theta}}(\mathbf{x}_t,t) \big )}$
    \State $\mathcal{L}_{\mathrm{FLM}} = -\frac{1}{L} \sum_l (\mathbf{x}_1^l)^\top \log \mathbf{p}_\theta^l$
    \State $\textcolor{plum}{\mathcal{L}_{\mathrm{Distill}} = - \frac{1}{L}\sum_l (\mathbf{q}_{\bar{\theta}}^l)^\top \log \mathbf{p}_\theta^l}$
    \State Update $\theta$: $\mathcal{L}_{\mathrm{SERF}} = \mathcal{L}_\mathrm{FLM} + \textcolor{plum}{\lambda \mathcal{L}_\mathrm{Distill}}$

    \State \textcolor{plum}{Update $\bar{\theta} \leftarrow \alpha \bar{\theta} + (1-\alpha) \theta$}
\Until{converged}
\end{algorithmic}
\end{algorithm}

\begin{algorithm}
\caption{\texttt{SERF-P}$_k$ Training (Probability Distribution Distillation from block $k$)}
\label{algo:serf_prob_layer}
\small

\begin{algorithmic}
\Require Dataset $\mathcal{D}$, $\lambda$, $\alpha$, \textcolor{plum}{block $k$}
\State \textbf{Initialize:} Student and Teacher $f_\theta$, $f_{\bar{\theta}}$, $\bar{\theta} \leftarrow \theta$, \textcolor{plum}{projection head $g_\theta$}
\Repeat
    \State $\mathbf{x}_1 \leftarrow \mathrm{onehot}(\mathbf{y}),\ \mathbf{y} \sim \mathcal{D}$;\quad $\boldsymbol{\epsilon} \sim \mathrm{N}(0, I)$
    \State Sample $s$, $\textcolor{plum}{t}$ \hfill $\triangleright$ See \autoref{sec:noise_lvl}
    \State $\mathbf{x}_s \leftarrow s\mathbf{x}_1 + (1-s)\boldsymbol{\epsilon}, \; \; \textcolor{plum}{\mathbf{x}_t \leftarrow t\mathbf{x}_1 + (1-t)\boldsymbol{\epsilon}}$
    \State $\mathbf{p}_\theta \leftarrow f_\theta(\mathbf{x}_s,s)$
    \State \textcolor{plum}{$\mathbf{p}^{(k)}_\theta \leftarrow o_{\bar{\theta}} \Big(g_\theta \big(\mathbf{h}_\theta^{(k)}(\mathbf{x}_s,s) \big)\Big), \;\; \mathbf{q}_{\bar{\theta}} \leftarrow \operatorname{sg}\big(f_{\bar{\theta}}(\mathbf{x}_t,t)\big)$}
    \State $\mathcal{L}_{\mathrm{FLM}} = - \frac{1}{L} \sum_l (\mathbf{x}_1^l)^\top \log \mathbf{p}_\theta^l$
    \State $\textcolor{plum}{\mathcal{L}_{\mathrm{Distill}} = - \frac{1}{L}\sum_l (\mathbf{q}_{\bar{\theta}}^l)^\top \log \mathbf{p}_\theta^{(k),l}}$
    \State Update $\theta$: $\mathcal{L}_{\mathrm{SERF}} = \mathcal{L}_\mathrm{FLM} + \textcolor{plum}{\lambda \mathcal{L}_\mathrm{Distill}}$
    \State \textcolor{plum}{Update $\bar{\theta} \leftarrow \alpha \bar{\theta} + (1-\alpha) \theta$}
\Until{converged}
\end{algorithmic}
\end{algorithm}

\begin{figure}[t]
    \centering
    \includegraphics[width=\textwidth]{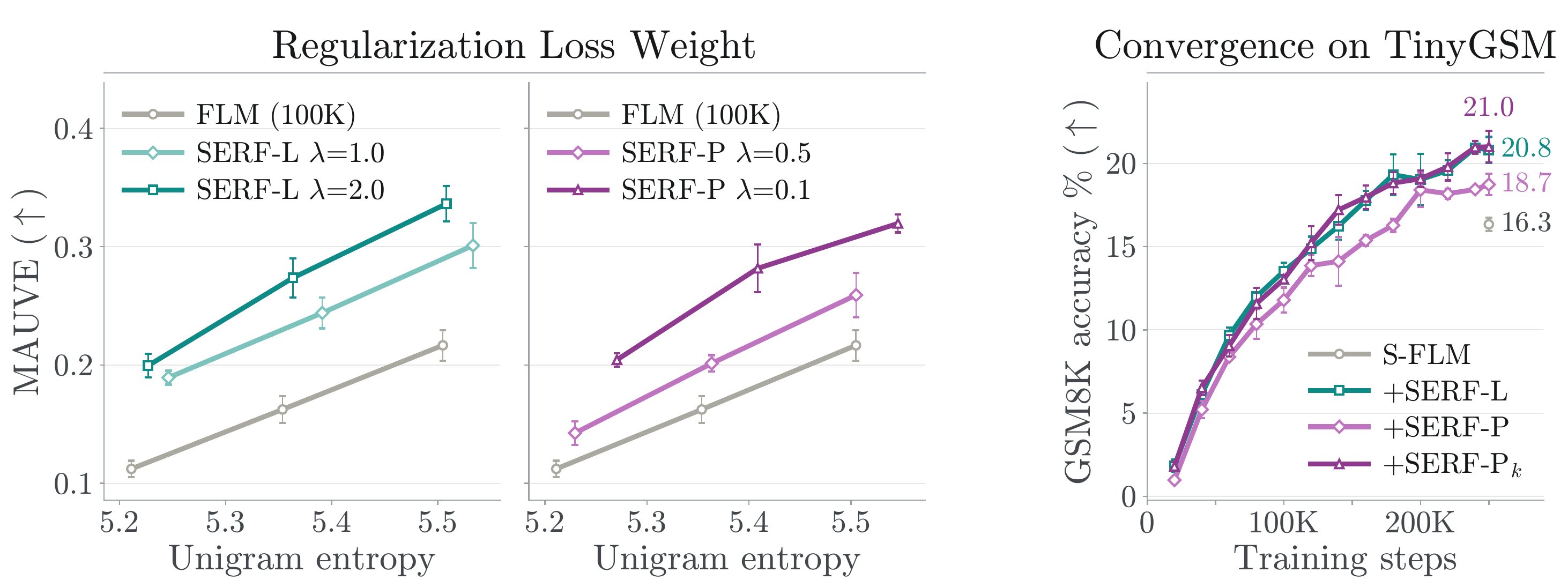}
\caption{\textbf{Left:}
Regularization weight ablation for \texttt{SERF-L} and \texttt{SERF-P}.
 \textbf{Right:} Training convergence of \texttt{SERF-L},
\texttt{SERF-P} and \texttt{SERF-P}$_k$  against the \texttt{S-FLM} baseline, trained on
\texttt{TinyGSM} and evaluated by \texttt{GSM8K} accuracy with $32$ sampling
steps.}

    \label{fig:ap_additional_results}
\end{figure}

\paragraph{Computational overhead.}
\texttt{SERF} adds no cost at sampling time. The teacher and the \texttt{SERF-L} projection head are
used only during training, and the model we sample from is the same network as the baseline. During training, each step needs one extra forward pass of the EMA teacher on
the cleaner input. This pass runs without gradients and keeps no activations for
the backward pass. A forward pass costs
roughly a third of a forward-backward step, so SERF makes each training step $\sim30\%$ slower as measured on our resources. The remaining additions (the EMA update of the teacher weights,
the small projection head and the \texttt{SERF} loss) are negligible. Even accounting for this per-step
overhead, $\texttt{SERF-L}$ reaches the quality of the baselines in much less wall-clock time, namely
7.7$\times$ faster for MAUVE, and Gen.~PPL, and 1.4$\times$ for \texttt{GSM8K} accuracy.

\paragraph{GSM8K evaluation.} We evaluate on all  \texttt{GSM8K} test
problems. The problem statement is supplied as a clean prefix and is held fixed
throughout sampling, so generation acts only on the answer.  Since this sampler is deterministic given the initial noise, we repeat
each evaluation over $5$ noise seeds and report the mean.

\paragraph{Baselines.} Baseline results for \texttt{FLM} and \texttt{S-FLM} are taken from their
respective papers. On \texttt{GSM8K} (\autoref{fig:gsm_results}), we compare
against two discrete diffusion language models,
\texttt{MDLM}~\citep{sahoo2024simple} and \texttt{DUO}~\citep{sahoo2025diffusion},
and three continuous diffusion language models,
\texttt{CANDI}~\citep{pynadath2025candi}, \texttt{FLM}~\citep{lee2026flow}
and \texttt{S-FLM}~\citep{deschenaux2026language}. All results in this table,
including the autoregressive baselines, are taken from
\citet{deschenaux2026language}.

\paragraph{Computational Resources.} For both training and sampling we use 8 NVIDIA GH200 $96$GB GPUs.

\subsection{Additional Evaluation Details} \label{sec:additiona_eval_details}

\paragraph{Entropy-matched generative perplexity.}
Generative perplexity can be lowered simply by sampling at a lower temperature,
which also lowers sample diversity. Comparing models at a fixed temperature is
therefore misleading, so we report Gen.\ PPL at a fixed sample entropy instead.
For each model  we draw
5{,}000 samples at a small sweep of temperatures, $\tau \in \{0.95, 1.0, 1.05\}$.
For each temperature we record the Gen.\ PPL and the mean
per-sequence unigram entropy $H$. We then interpolate the sweep to a target
entropy $H = 5.44$, which is the entropy of OpenWebText validation text.

\section{Additional Experimental Results}

\label{app:additional_results}

\subsection{Training acceleration on TinyGSM}

Figure~\ref{fig:ap_additional_results} (right) tracks \texttt{GSM8K} accuracy over training, measured with
32 sampling steps and averaged over 4 evaluation seeds. Both \texttt{SERF} variants learn
faster than the \texttt{S-FLM} baseline. SERF-L reaches the accuracy of the released S-FLM
checkpoint, trained for 250K steps (16.3\%), after only 140K steps (16.2\%), that is,
with $1.8\times$ fewer training steps. \texttt{SERF-P} does so after 180K steps
($1.4\times$ fewer). At 250K steps, SERF-L reaches
20.8\% and SERF-P 18.7\%, respectively 4.5 and 2.4 points above S-FLM. Taking into account SERF's per-step overhead of roughly 30\%, SERF-L still matches
the baseline $1.4\times$ faster in training time.

\vspace{-10pt}
\subsection{Effect of the self-regularization weight $\lambda$.}
\autoref{fig:ap_additional_results} (left) reports MAUVE against unigram entropy for
\texttt{SERF-L} and \texttt{SERF-P} under different regularization weights.
The two variants respond to $\lambda$ in opposite directions: \texttt{SERF-L}
benefits from a larger weight ($0.336$ at $\lambda=2.0$ vs.\ $0.321$ at
$\lambda=1.0$), whereas \texttt{SERF-P} degrades as the weight grows ($0.320$
at $\lambda=0.1$ vs.\ $0.259$ at $\lambda=0.5$). This reflects the scale of
the two alignment losses. The cosine distance used by \texttt{SERF-L} is
bounded and much smaller in magnitude than the cross-entropy of
$\mathcal{L}_{\mathrm{FLM}}$, so a larger $\lambda$ is needed for it to
meaningfully affect training. In contrast, \texttt{SERF-P} employs a
cross-entropy loss on the same scale as $\mathcal{L}_{\mathrm{FLM}}$, thus increasing $\lambda$ shifts the
effective target away from the ground truth and toward the
teacher's prediction.

\section{Qualitative Comparisons}
\label{sec:app_qual}
\begin{figure}[t]
    \centering
    \includegraphics[width=0.9\textwidth]{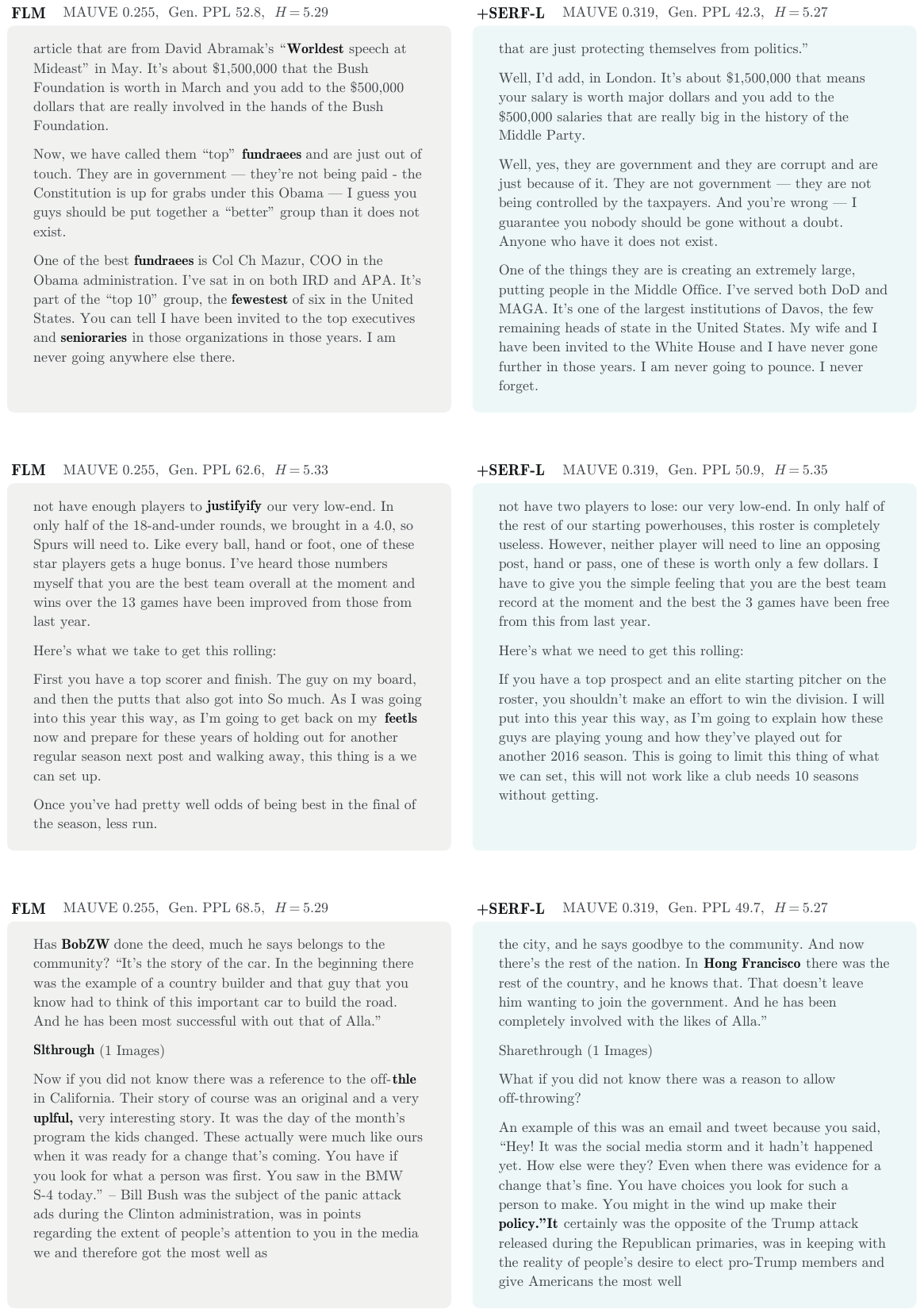}
\caption{\textbf{Generated samples on \texttt{OWT}.} Samples from
\texttt{FLM} (left) and \texttt{SERF-L} (right), all trained
for $1$M and $300K$ steps respectively. Text generated with $1024$ sampling steps from the same
initial noise. Errors are highlighted in \textbf{bold}.}

    \label{fig:ap_text_example_1}
\end{figure}

\begin{figure}[t]
    \centering
    \includegraphics[width=0.9\textwidth]{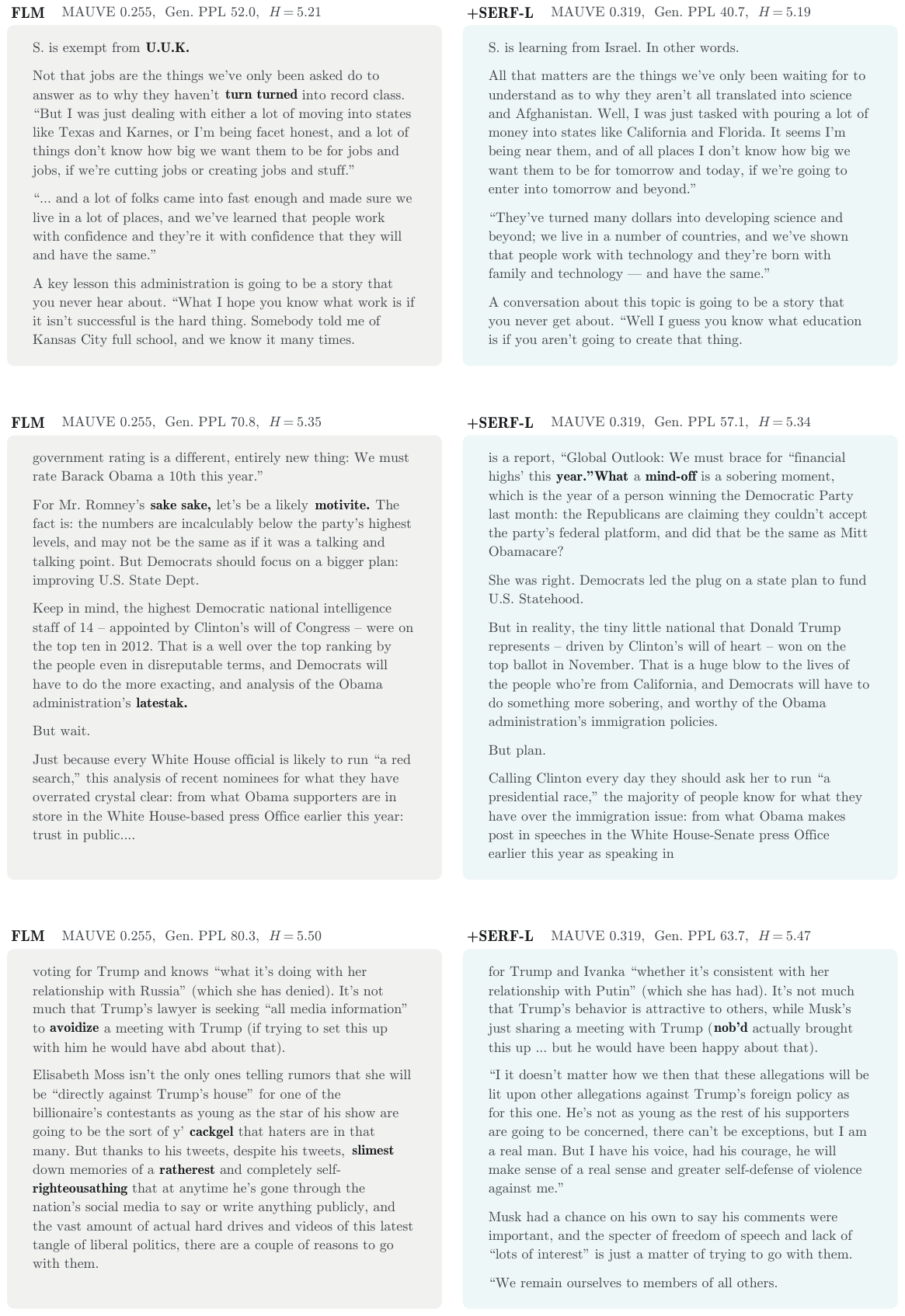}
\caption{\textbf{Generated samples on \texttt{OWT}.} Samples from
\texttt{FLM} (left) and \texttt{SERF-L} (right), all trained
for $1$M and $300K$ steps respectively. Text generated with $1024$ sampling steps from the same
initial noise. Errors are highlighted in \textbf{bold}.}

    \label{fig:ap_text_example_2}
\end{figure}

\begin{figure}[t]
    \centering
    \includegraphics[width=0.9\textwidth]{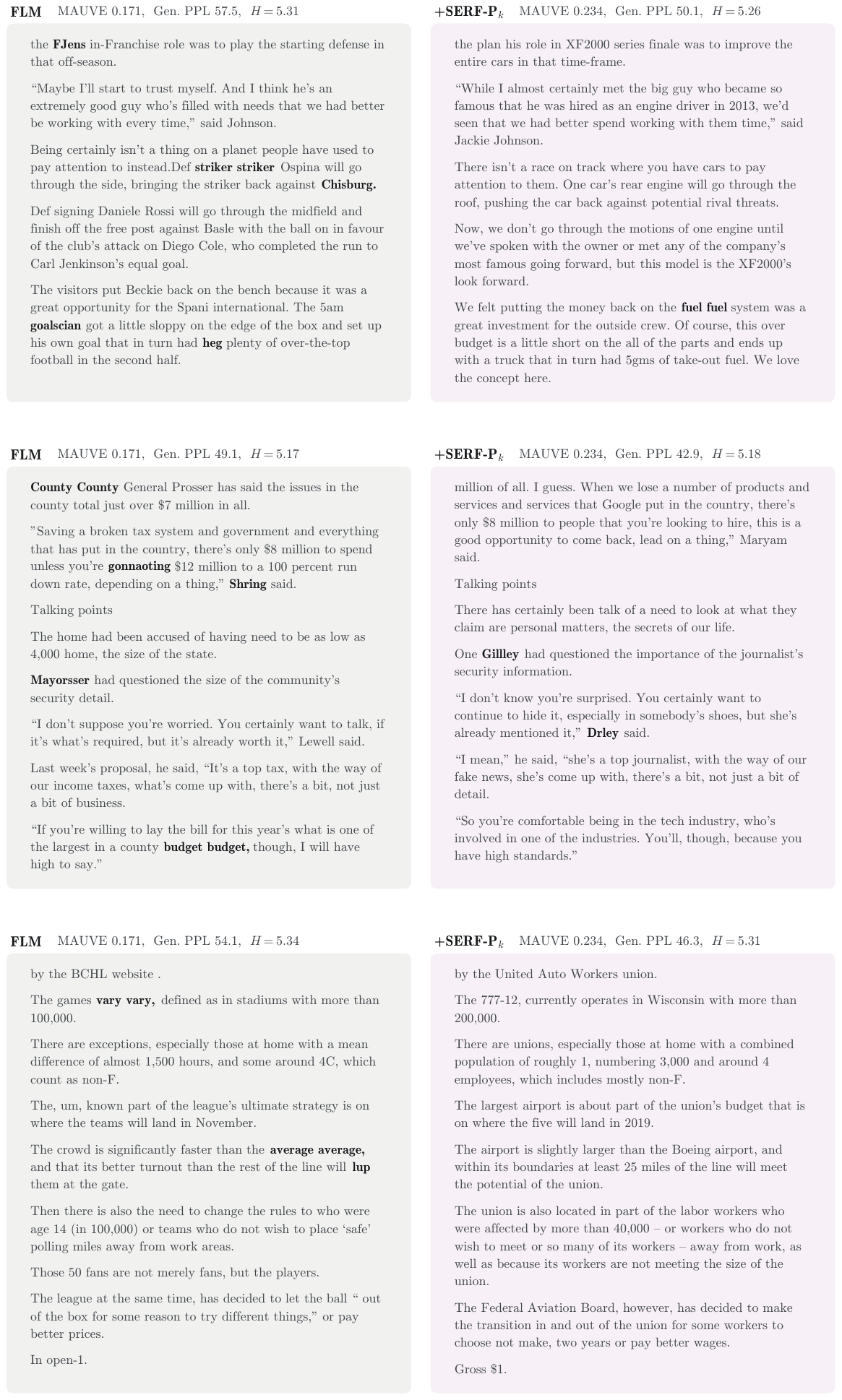}
\caption{\textbf{Generated samples on \texttt{OWT}.} Samples from
\texttt{FLM} (left) and \texttt{SERF-P}$_k$ (right), all trained
for $1$M and $300K$ steps respectively. Text generated with $1024$ sampling steps from the same
initial noise. Errors are highlighted in \textbf{bold}.}

    \label{fig:ap_text_example_1_serp}
\end{figure}

\begin{figure}[t]
    \centering
    \includegraphics[width=0.9\textwidth]{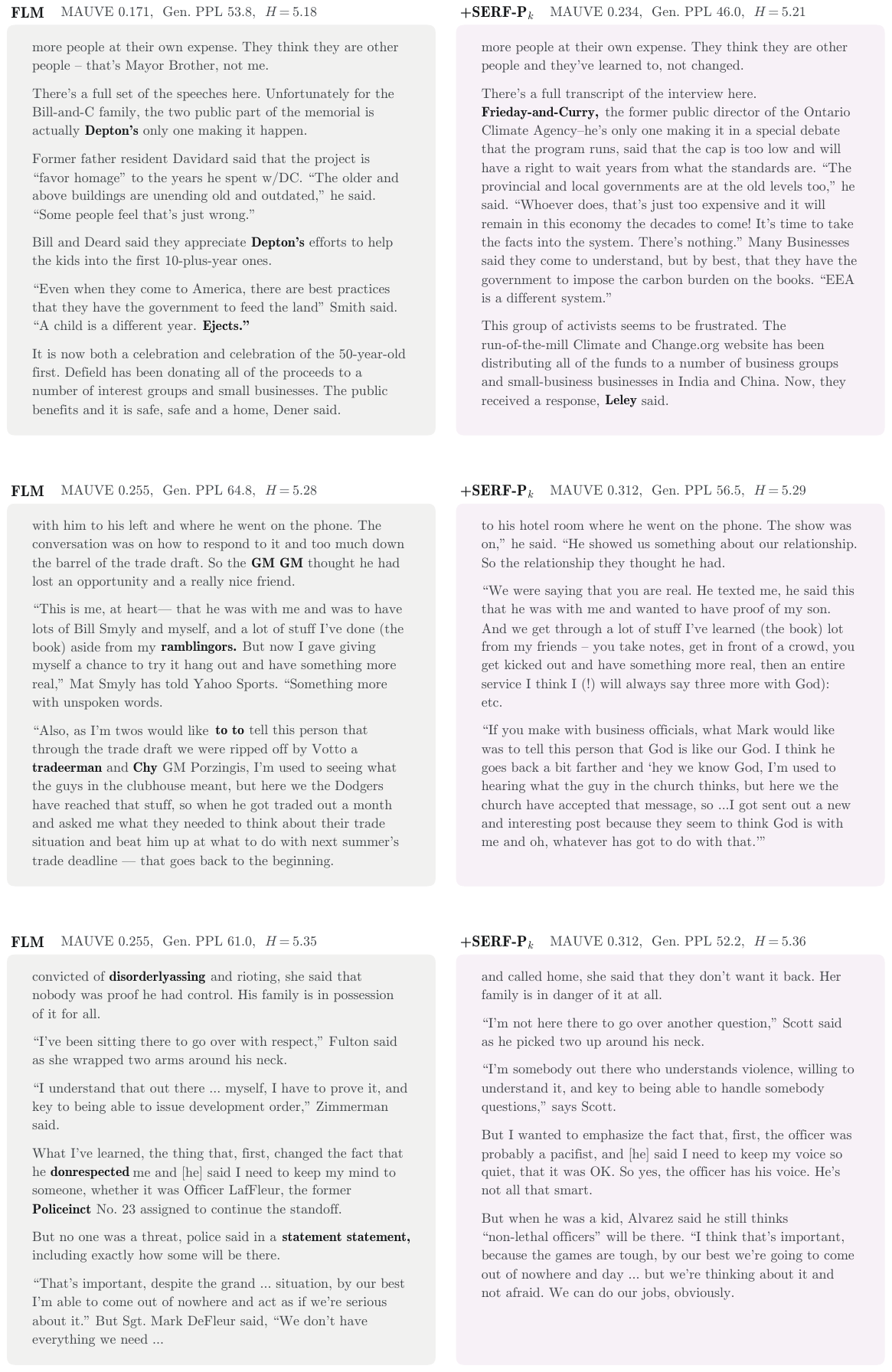}
\caption{\textbf{Generated samples on \texttt{OWT}.} Samples from
\texttt{FLM} (left) and \texttt{SERF-P}$_k$ (right), all trained
for $1$M and $300K$ steps respectively. Text generated with $1024$ sampling steps from the same
initial noise. Errors are highlighted in \textbf{bold}.}

    \label{fig:ap_text_example_2_serp}
\end{figure}

We present qualitative comparisons with the baseline \texttt{FLM} in
\autoref{fig:ap_text_example_1}, \autoref{fig:ap_text_example_2}, \autoref{fig:ap_text_example_1_serp} and \autoref{fig:ap_text_example_2_serp}. For each comparison, we use the same initial Gaussian noise and
vary only the trained denoiser, generating samples with 1024 sampling steps. Since Euler sampling is deterministic given the initial
noise, this setup allows a direct comparison between the two learned flows. As
shown in the examples, \texttt{SERF} produces more coherent and grammatically
well-formed text. In contrast, the baseline \texttt{FLM} more often generates degenerate
fragments, repeated symbols, and locally incoherent phrases.

\section{Limitations and Future Work}
\label{sec:limitations}
Our method is currently developed and evaluated for continuous flow-based language models. Extending the framework to discrete diffusion language models \citep{sahoo2024simple, schiff2024simple} represents a promising direction for future work. Since discrete diffusion models follow different corruption and denoising dynamics than continuous flows, adapting the teacher supervision mechanism may require additional design considerations. We view this as an opportunity to further generalize our framework and broaden its applicability.

\begin{table}[t]
\centering
\caption{\textbf{Sources of the numbers in Fig.~\ref{fig:teaser}.} For each value, we list
the compared models and where the underlying numbers are reported.}
\label{tab:fig1_sources}
\small
\setlength{\tabcolsep}{6pt}
\renewcommand{\arraystretch}{1.15}
\begin{tabular}{@{}l r l l@{}}
\toprule
\Th{Metric} & \Th{Value} & \Th{Comparison} & \Th{Source} \\
\midrule
\rowcolor{gray!10} \multicolumn{4}{@{}l}{\textit{Left: wall-clock training acceleration}} \\
MAUVE      & $\sim 7.7\times$ & \texttt{SERF-L} vs.\ \texttt{FLM}
           & ~\autoref{fig:various} (left) \\
Gen.\ PPL  & $\sim 7.7\times$ & \texttt{SERF-L} vs.\ \texttt{FLM}
           & ~\autoref{fig:various} (right) \\

GSM8K Acc. & $\sim 1.4\times$ & \texttt{SERF-L} vs.\ \texttt{S-FLM}
           & ~\autoref{fig:ap_additional_results} (right), \\
\midrule
\rowcolor{gray!10} \multicolumn{4}{@{}l}{\textit{Middle: relative improvement at the same sampling budget}} \\
MAUVE      & $+24.3\%$ & \texttt{SERF-L} vs.\ \texttt{FLM}
           & ~\autoref{tab:best_mauve_steps} (top) \\
Gen.\ PPL  & $+2.8\%$ & \texttt{SERF-L} vs.\ \texttt{FLM}
           & ~\autoref{tab:best_mauve_steps} (bottom) \\
GSM8K Acc. & $+28.8\%$ & \texttt{SERF-P}$_k$ vs.\ \texttt{S-FLM}
           & ~\autoref{fig:gsm_results} (left) \\
\midrule
\rowcolor{gray!10} \multicolumn{4}{@{}l}{\textit{Right: sampling acceleration}} \\
MAUVE      & $\sim 2\times$  & \texttt{SERF} vs.\ \texttt{FLM}
           & ~\autoref{tab:best_mauve_steps} (top) \\
Gen.\ PPL  & $\sim 4\times$  & \texttt{SERF} vs.\ \texttt{FLM}
           & ~\autoref{tab:best_mauve_steps} (bottom) \\
GSM8K Acc. & $\sim 64\times$ & \texttt{SERF} vs.\ \texttt{S-FLM}$^{*}$
           & ~\autoref{fig:gsm_results} (middle) \\
\bottomrule
\end{tabular}
\end{table}

\end{document}